\documentclass[final]{cvpr}

\usepackage{times}
\usepackage{epsfig}
\usepackage{graphicx}
\usepackage{amsmath}
\usepackage{amssymb}
\usepackage{bbding}

\usepackage{acro}
\usepackage{ctable}
\usepackage{multirow}
\usepackage{amsmath}

\DeclareMathOperator*{\argmin}{arg\,min}

\usepackage[pagebackref=true,breaklinks=true,colorlinks,bookmarks=false]{hyperref}

\DeclareAcronym{CT}{short=CT, long=computed tomography}
\DeclareAcronym{Dense}{short=DenseNet, long=dense convolutional network}
\DeclareAcronym{CNN}{short=CNN, long=convolutional neural network}
\DeclareAcronym{SSL}{short=SSL, long=self-supervised learning}
\DeclareAcronym{RECIST}{short=RECIST, long=response evaluation criteria in solid tumours}
\DeclareAcronym{anatomy_net}{short=ASE, long=anatomy signal encoder}
\DeclareAcronym{my_speedup}{short=FCC, long=fast cross-correlation}
\DeclareAcronym{my_method}{short=DLT, long=deep lesion tracker}
\DeclareAcronym{my_dataset}{short=DLS, long=deep longitudinal study} 

\def\Fig#1{{Fig.~\ref{fig:#1}}}
\def\Eq#1{{Eq.~\ref{eq:#1}}}
\def\Table#1{{Table~\ref{tbl:#1}}}
\def\Sec#1{{Sec.~\ref{sec:#1}}}

\begin{document}

\title{Deep Lesion Tracker: Monitoring Lesions in 4D Longitudinal Imaging Studies}
\author{Jinzheng Cai$^1$, Youbao Tang$^1$, Ke Yan$^1$, Adam P. Harrison$^1$, Jing Xiao$^2$, Gigin Lin$^3$, Le Lu$^1$ \\
$^1$ PAII Inc., Bethesda, MD, USA ~~~~ $^2$ Ping An Technology, Shenzhen, PRC \\ 
$^3$ Chang Gung Memorial Hospital, Linkou, Taiwan, ROC \\
{\tt\small caijinzhengcn@gmail.com, tiger.lelu@gmail.com}
}
\maketitle

\begin{abstract}
   Monitoring treatment response in longitudinal studies plays an important role in clinical practice. Accurately identifying lesions across serial imaging follow-up is the core to the monitoring procedure. Typically this incorporates both image and anatomical considerations. However, matching lesions manually is labor-intensive and time-consuming. In this work, we present \ac{my_method}, a deep learning approach that uses both appearance- and anatomical-based signals. To incorporate anatomical constraints, we propose an anatomical signal encoder, which prevents lesions being matched with visually similar but spurious regions. In addition, we present a new formulation for Siamese networks that avoids the heavy computational loads of 3D cross-correlation. To present our network with greater varieties of images, we also propose a \ac{SSL} strategy to train trackers with unpaired images, overcoming barriers to data collection. To train and evaluate our tracker, we introduce and release the first lesion tracking benchmark, consisting of $3891$ lesion pairs from the public DeepLesion database. The proposed method, \ac{my_method}, locates lesion centers with a mean error distance of 7$mm$. This is 5\% better than a leading registration algorithm while running 14 times faster on whole CT volumes. We demonstrate even greater improvements over detector or similarity-learning alternatives. \ac{my_method} also generalizes well on an external clinical test set of $100$ longitudinal studies, achieving 88\% accuracy. Finally, we plug \ac{my_method} into an automatic tumor monitoring workflow where it leads to an accuracy of 85\% in assessing lesion treatment responses, which is only 0.46\% lower than the accuracy of manual inputs. We release our benchmark at \url{https://github.com/JimmyCai91/DLT}.
\end{abstract}

\acresetall

\section{Introduction}
\begin{figure}
    \centering
    \includegraphics[width=\linewidth]{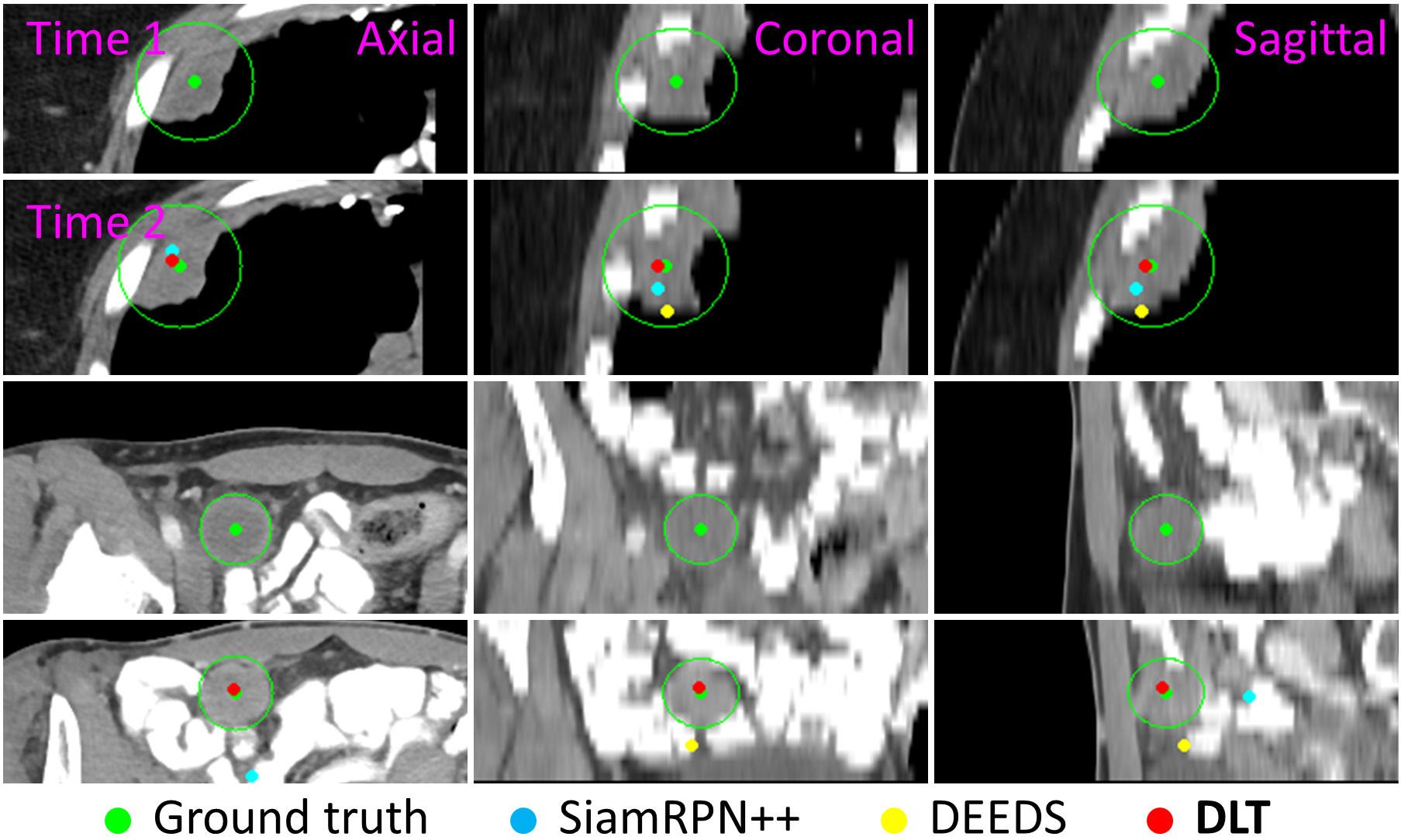}
    \caption{Comparison of our approach with two state-of-the-art approaches for 3D tracking. The proposed \ac{my_method} can predict lesion centers more precisely than SiamRPN++ \cite{Li2019SiamRPN++} and DEEDS \cite{Heinrich2013DeedsBCV}.}
    \label{fig:examples} 
\end{figure}
Monitoring treatment response by identifying and measuring corresponding lesions is critical in radiological workflows \cite{Eisenhauer2009RECIST,RafaelPalou2020Noudle,Ardila2019LungCancer,Tan2016OvarianCancer}. Manually conducting these procedures is labor-intensive, as expert clinicians must review multiple images and go back and forth between these images for comparison. This is usually subject to considerable inter-observer variability \cite{tang2018semi}. Therefore, computer aided tools have the opportunity to lower costs, increase turnaround speeds, and improve reliability.

Automatic image-based lesion monitoring can be decomposed into several sub-procedures: (1) detect lesions of interest; (2) then track instances of the same lesion across different time points; and (3) measure changes among the identified instances. The first step of detecting lesions of interest can be formulated as object detection. In general, the computer vision field has made progress toward this problem \cite{Girshick2014RCNN,Lin2020FocalLoss,Zhou2019CenterNet}. However, medical imaging has its distinct challenges as the data is often in 3D format, \eg{}, \ac{CT}, and usually the required annotations are unavailable. Therefore, there are efforts to improve object detection with medical images \cite{Jiang2020ElixirNet,Shao2019Attentive,Wang2019UOD,Cai2020VULD,Yan2020Lens,Tang2019Uldor,Roth2014LNDetection}. Similarly step (3) also has many viable solutions because it can be formulated as (3D) object segmentation, which is a fundamental topic that attracts attentions from both computer vision \cite{Long2015FCN,Xie2017HED,Chen2018DeepLab} and medical image analysis \cite{Ronneberger2015UNet,Milletari2016VNet,Cicek2016UNet3D,Roth2015DeepOrgan,Cai2018SlicePropagation,Tang2020OneClick}. In contrast, step (2), tracking the same lesion across different time points, is not as well developed as lesion detection and segmentation. Part of the lack of development can be attributed to the lack of good benchmark datasets to evaluate performance. In this work, we address this by both introducing a public benchmark and also formulating a powerful lesion tracking solution, called \ac{my_method}, that can accurately match instances of the same lesion across different images captured at different time points and contrast phases by using both appearance and anatomical signals. In \Fig{examples}, we show two real-life examples of lesion tracking.

Similar with visual tracking in the general computer vision, lesion tracking can be viewed as to match instances of the same lesion in neighboring time frames. However, it is challenging due to changes in size and appearance. Lesion size can enlarge multiple times than its baseline or nadir. Meanwhile, its appearance varies during the follow-up exam because of morphological or functional changes, commonly attributed to necrosis or changes in vascularity. Therefore, an effective tracker should handle both size and visual changes of lesions. Trackers based on image registration \cite{Ardila2019LungCancer,Tan2016OvarianCancer} are robust to appearance changes, as registration inherently introduces anatomical constraints for lesion matching. The involved body part and surrounding organs of the target lesion are constrained among different images. However, registration algorithms \cite{Heinrich2013DeedsBCV,Heinrich2015DeedsBCV,Marstal2016SimpleElastix,Miao2016Registration,Balakrishnan2019VoxelMorph} are usually less sensitive to local image changes; thus, they can be inaccurate to track small-sized lesions or lesions with large shape changes. On the other hand, appearance-based trackers \cite{RafaelPalou2020Noudle,Gomariz2019LiverLandmark} handle size and appearance changes by projecting lesion images into an embedding space \cite{Yan2018LesionGraph,Yan2019LesaNet}, where images of the same lesion have similar embeddings and images of different lesions are different from one another. However, these appearance-based trackers may mismatch lesions with visually similar but spurious backgrounds. Therefore, to combine the merits of both strategies, we design our tracker to conduct appearance based recognition under anatomical constraints. 

Because the proposed \acf{my_method} is a deep learning model, providing enough training data is a prerequisite for good performance. To this end, we construct a dataset with 3891 lesion pairs, collected from DeepLesion \cite{Yan2018DeepLesion}, to train and evaluate different tracking solutions. We publicly release the annotations to facilitate related research\footnote{\url{https://github.com/JimmyCai91/DLT}}. Although more training pairs can promote a stronger tracker, labor and time costs preclude easily collecting and annotating a large number of longitudinal studies for a specific clinical application. Therefore, we also introduce an effective \acf{SSL} strategy to train trackers. Importantly, this strategy can train lesion trackers using images from only one time point, meaning non-longitudinal datasets can be used, which are more readily collected. This allows for a more ready introduction of more lesion instances with varied appearances and sizes. 

With the proposed \ac{my_method} and model training strategies, we achieve 89\% matching accuracy on a test set of 480 lesion pairs. Meanwhile, we demonstrate that \ac{my_method} is robust to inaccurate tracking initializations, \ie{}, the given initial lesion center. In our robustness study, inaccurate initialization causes 10\% accuracy drops on SiamRPN++ \cite{Li2019SiamRPN++} and DEEDS \cite{Heinrich2013DeedsBCV}. In contrast, the accuracy of \ac{my_method} only decreases by 1.9\%. We then apply \ac{my_method} to an external testing set of 100 real-life clinical longitudinal studies, delivering 88\% matching accuracy and demonstrating excellent generalizability. Finally, we plug \ac{my_method} into a lesion monitoring pipeline to simulate automatic treatment monitoring. The workflow assesses lesion treatment responses with 85\% accuracy, which is only 0.46\% lower than the accuracy of manual inputs.

\section{Related Work}
Visual object tracking is an active research topic in general computer vision \cite{Bolme2010MOSSE,Ma2015CF2,Danelljan2015DeepSRDCF,Wang2015SODLT,Wang2015FCNT,Bertinetto2016STAPLE,Nam2016MDNet,Tao2016SINT,Valmadre2017CFNet}.
We focus our review on recent progresses, especially deep learning based approaches.

{\bf Tracking as Similarity Learning.} Tracking of target objects can be achieved via similarity comparisons between the object template and proposals from the search domain. Similarities are measured by either color/intensity representations \cite{Danelljan2014DSST}, spatial configurations \cite{Yao2013PBV,Liu2015PBV}, or their combinations \cite{Bertinetto2016STAPLE}. Recently, deep learning features are more widely used for visual tracking \cite{Wang2015FCNT,Nam2016MDNet,Danelljan2015DeepSRDCF,Wang2015SODLT} as they outperform hand-crafted features with more expressive representations. To efficiently extract and compare deep learning features, SiamFC \cite{Bertinetto2016SiameseFC} and CFNet \cite{Valmadre2017CFNet} use a cross-correlation layer at the end of Siamese architectures \cite{Bromley1993Siamese}. This cross-correlation layer uses Siamese feature maps extracted from the template image patch as a kernel to operate fully circular convolution on the corresponding Siamese feature maps of the search image. This procedure encodes the information regarding the relative position of the target object inside the search image. Within the same framework of SiamFC, SiamRPN++ \cite{Li2019SiamRPN++} introduced strategies to allow training of Siamese networks with modern very deep networks, \eg{}, \ac{Dense} \cite{Huang2017DenseNet}, to further boost tracking accuracy. This is critical for medical image analysis as many medical applications lack large-scale training data and rely on transfer learning of pre-trained networks for good performance \cite{Shin2016CNN}.

Siamese networks have also been investigated in medical image analysis. Gomariz \etal{} \cite{Gomariz2019LiverLandmark} applied 2D Siamese networks to track liver landmarks in ultra-sound videos. Liu \etal{} \cite{Liu2020COSD-CNN} extended similar 2D Siamese networks in a coarse-to-fine fashion. While, Rafael-Palou \etal{} \cite{RafaelPalou2020Noudle} performed 3D Siamese networks with \ac{CT} series, only shallow network architectures were evaluated on tracking lung nodules. However, we follow SiamRPN++ \cite{Li2019SiamRPN++} to use Siamese networks with 3D \ac{Dense} backbones and apply it to conduct universal lesion tracking in whole body \ac{CT} images. Processing different types of lesions with a unified deep learning model \cite{Shao2019Attentive,Tang2019Uldor,Yan2018DeepLesion,Yan2018LesionGraph,Yan2019LesaNet,Yan2019Mulan,Cai2020LesionHarvester,Cai2020VULD} demonstrates computational efficiency and could also alleviate model over-fitting. Different from prior formulations of Siamese networks, we propose a simple but effective 3D kernel decomposition to speed up 3D cross-correlation operations for object matching. This provides dramatic boosts in efficiency, reducing over 65\% of FLOPs in our \ac{my_speedup} layer. 

\begin{figure*}[ht]
    \centering 
    \includegraphics[width=0.98\linewidth]{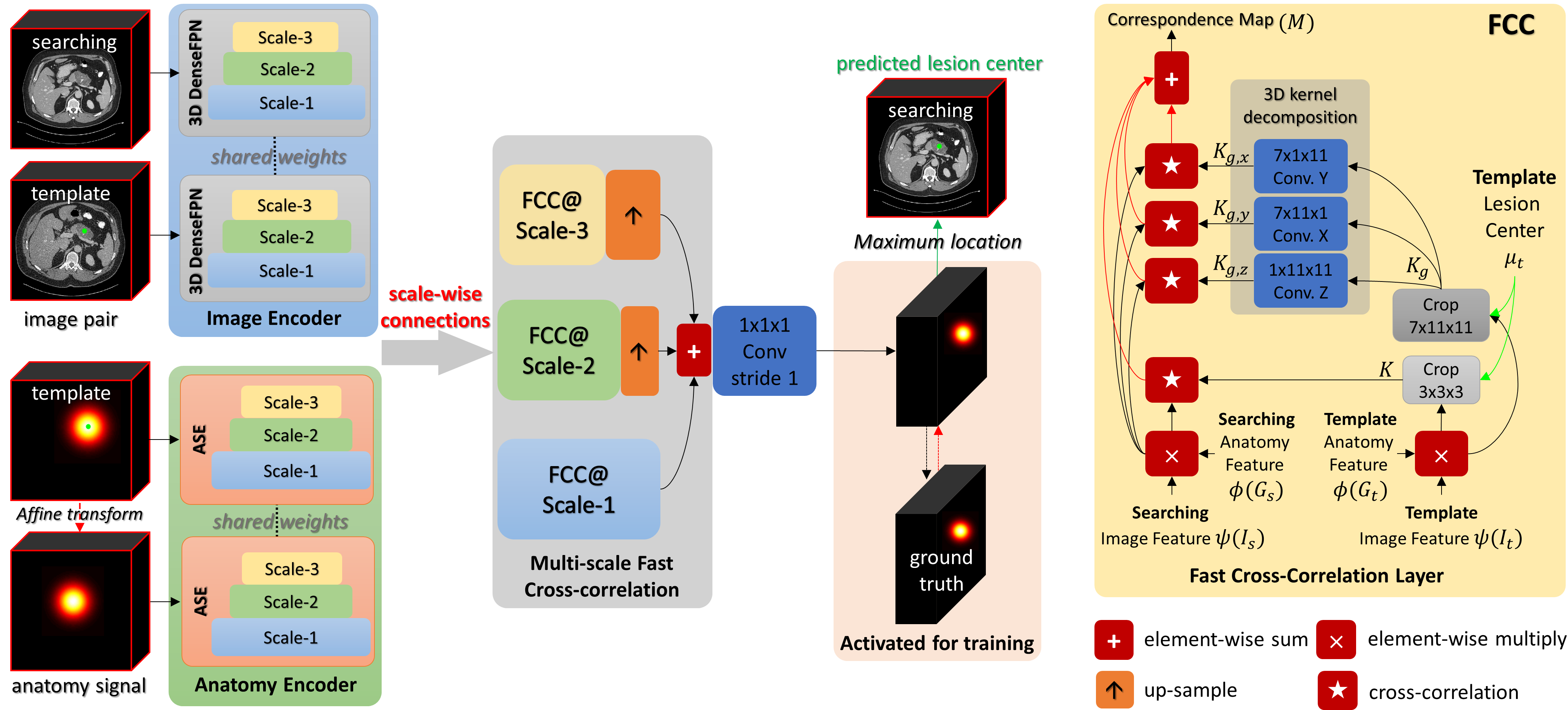}
    \caption{The configuration of our proposed \acl{my_method}.}
    \label{fig:vlt}
\end{figure*}

{\bf Tracking as Detector Learning.} Tracking as detector learning relies on developing discriminative models to separate the target from background regions \cite{Avidan2004SVT,Babenko2011TrackingMIL,Henriques2012Circulant,Danelljan2014DSST}. A discriminative model that is suitable for visual tracking should consist of two core components, namely a classifier that can be efficiently updated online during visual tracking \cite{Avidan2004SVT,Babenko2011TrackingMIL,Henriques2012Circulant} and a powerful feature representation, \eg{} features extracted by \acp{CNN} \cite{Krizhevsky2012AlexNet,Huang2017DenseNet} that can let the classifier easily differentiate objects in the feature space. Following this strategy, SO-DLT \cite{Wang2015SODLT}, FCNT \cite{Wang2015FCNT}, and MDNet \cite{Nam2016MDNet} all train \acp{CNN} offline from large-scale object recognition tasks so that the learnt feature representation is general with visual objects. During tracking, they freeze the lower layers of the network as a feature extractor and update the higher layers to adapt to the specific video domain.

In this work, we consider the strategy of tracking via detector learning and accordingly construct our strong lesion tracking baselines. Given the specialty of processing medical data, especially 4D \ac{CT} images (3D image plus time), there are no baseline methods ready for comparison. Thus, we construct our own lesion tracking baselines by concatenating the state-of-the-art lesion detection \cite{Cai2020VULD,Yan2020Lens} models with deep learning feature extractors \cite{Yan2018LesionGraph,Yan2019LesaNet}. However, the tracker developed with this strategy can be sub-optimal since the detection models and feature extractors are developed from independent offline tasks. In contrast, our proposed \ac{my_method} unifies the tasks of feature extraction and target object localization in an end-to-end structure and outperforms these detector learning baselines with higher accuracy and faster speed.

{\bf Tracking Priors from Image Registration.} Visual tracking in video follows a prior of spatial consistency, which means the search space in the next video frame can be constrained to be near to the current location. This prior is helpful for improving tracking efficiency and making the model robust to background distractors \cite{Tao2016SINT,Bertinetto2016SiameseFC,Gomariz2019LiverLandmark}. Similarly, lesion tracking in \ac{CT} should follow a spatial consistency governed by anatomical considerations. This implies that the surrounding organs and structures of a lesion will not drastically change. Under such constraints, image registration approaches \cite{Heinrich2013DeedsBCV,Heinrich2015DeedsBCV,Marstal2016SimpleElastix,Miao2016Registration,Balakrishnan2019VoxelMorph} can perform lesion tracking via image alignment. Specifically, registration algorithms are designed to optimize the global structural alignment, \ie{} accurately align boundaries of large organs, while being robust to local changes. Nonetheless, although reported results suggest that registration algorithms are useful for aligning large-sized lesions \cite{Tan2016OvarianCancer,Raju2020Liver,Zhang2020PDAC}, they can fail to track small-sized lesions and struggle whenever there are local changes in the lesion's appearance.  

In this work, we improve upon the capabilities of registration approaches using deep learning based lesion appearance recognition to match lesions based on both visual and anatomical signals. Specifically, we first roughly initialize the location of a target lesion using image registration, \ie{}, affine registration \cite{Marstal2016SimpleElastix}. Then, our proposed deep learning model, \ac{my_method}, refines the location to the lesion center using appearance-based cues. In contrast with approaches that use the spatial and structural priors simply in pre- \cite{Tao2016SINT,Bertinetto2016SiameseFC} or post-processing \cite{Gomariz2019LiverLandmark}, \ac{my_method} takes them as its inputs and propagates them together with \ac{CT}-based visual signal to generate the final target location. The priors also function as attention guidance, letting the appearance learning focus on vital image regions. 

\section{Deep Lesion Tracker}
We build \ac{my_method} based on the structure of Siamese networks  because they are efficient and deliver state-of-the-art visual tracking performance for many computer vision tasks. The core component of Siamese-based tracking is a correlation filter, which is also known as cross-correlation layer. It uses Siamese features extracted from the template image patch as a kernel to perform explicit convolutional scanning over the entire extent of the search image feature masps. \Fig{vlt} shows its overall configuration. Our goal is to apply the proposed model to process three dimensional medical data, \ie{}, \ac{CT} images. Therefore, we create network backbones in 3D and introduce an \acf{anatomy_net} to guide lesion tracking with anatomical constraints. To avoid the prohibitive computational expenses of 3D cross-correlation between the template and the search image, we introduce a simple but effective formulation to speed up this procedure.

{\bf Problem definition.} We use $I_t$ and $I_s$ to respectively denote a template and search \ac{CT} image. In $I_t$, a lesion is known with its center $\mu_t$ and radius $r_t$. Given $I_t$, $I_s$, $\mu_t$, and $r_t$, the task of lesion tracking is to locate the same lesion in $I_s$ by predicting its new center $\mu_s$.

\subsection{Image Encoder: 3D DenseFPN}
\begin{figure*}[t!]
   \centering
   \includegraphics[width=0.98\linewidth]{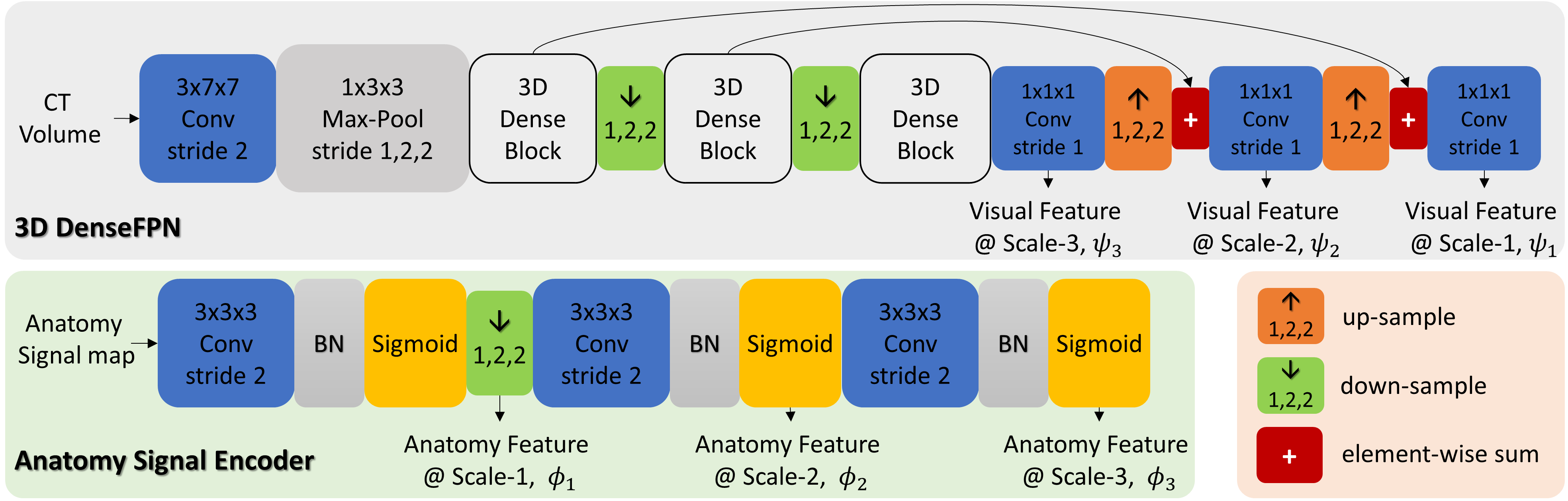}
   \caption{Network configurations of the proposed image encoder 3D DenseFPN and \acf{anatomy_net}.}
   \label{fig:networks}
\end{figure*}

In lesion tracking, the Siamese network needs to process lesions with varied appearances and sizes in 3D images. As shown in \Fig{networks}, we use a deep 3D image encoder with large model capacity, so that it can learn effective feature representations. Specifically, we transform \ac{Dense} into 3D by duplicating its 2D convolutional kernels along the third direction and then downscaling weight values by the number of duplications \cite{Carreira2017I3D}. This configuration is found to be more effective than 3D UNet \cite{Cicek2016UNet3D} on modeling universal lesion appearances \cite{Cai2020VULD}. We then add a feature pyramid network (FPN) \cite{Lin2017FPN} after the 3D \ac{Dense} to generate visual features at three scales. We visually depict the detailed configuration of 3D DenseFPN in \Fig{networks}. For clarity, we use $\psi_1$, $\psi_2$, and $\psi_3$ to refer to the image mapping functions that generate feature maps from the largest to the smallest resolutions, respectively.

\subsection{Anatomy Signal Encoder and Its Inputs} 

We observe that directly implementing lesion tracking with Siamese networks can produce matches with visually similar but spurious regions. 
In contrast, affine registration \cite{Marstal2016SimpleElastix} is a robust approach to roughly align \ac{CT} images. It is achieved by solving 
\begin{equation} \label{eq:aff}
    \mathcal{T}_{\text{Aff}} = \argmin_{\mathcal{T}_{\text{Aff}}\in \mathcal{A} } \|\mathcal{T}_{\text{Aff}}(I_t) - I_s\|_1, 
\end{equation}
where $\mathcal{A}$ is the space of affine transforms. The projected location of the template lesion, $\mathcal{T}_{\text{Aff}}(\mu_t)$, is usually located close to the actual target lesion.  While prior art has used affine registration as pre- \cite{Tao2016SINT,Bertinetto2016SiameseFC} or post-processing \cite{Gomariz2019LiverLandmark}, these do not provide mechanisms for incorporation into a tracking pipeline that cross-correlates template features across the entire extent of the search image. For example, pre-registering will have minimal effect on the translation-invariant  cross-correlation. Instead, as shown in \Fig{vlt}, we encode anatomy signals as Gaussian heatmaps centered at lesion locations: 
\begin{equation}\label{eq:ase_heatmap}
    \mathcal{G}(\mu, nr) = \exp\left(-\frac{\sum_{i\in\{x,y,z\}}{(i-\mu^i)^2}}{2(nr)^{2}}\right),
\end{equation}
where we find $n=4$ delivers the best performance. For $I_t$ we simply use the template lesion location and size: $\mathcal{G}(\mu_t, n r_t)$. For $I_s$ we use the affine-projected location and size of the template lesion: $\mathcal{G}(\mathcal{T}_{\text{Aff}}(\mu_t), n\mathcal{T}_{\text{Aff}}(r_t))$. For clarity, we simply refer to the template and search anatomy signal maps as $G_t$ and $G_s$, respectively. We solve \Eq{aff} using SimpleElastix \cite{Marstal2016SimpleElastix}.

\Fig{networks} depicts the network configuration of the proposed \ac{anatomy_net}. It encodes anatomical signals into high-dimensional anatomical features with three different resolutions. In correspondence with 3D DenseFPN, we denote the network functions for the three scales as $\phi_1$, $\phi_2$, and $\phi_3$ from the largest to the smallest, respectively. 

\subsection{Fast Cross-Correlation}

As mentioned, correlation is a core operation of Siamese-based tracking, which creates a correspondence map between target and search features, $\psi(I_t)$ and $\phi(G_t)$, respectively. Because we perform the same operation at each scale, we drop the scale subscripts here for simplicity. To conduct cross-correlation, we first fuse image and anatomy features.  For example, to fuse $\psi(I_t)$ and $\phi(G_t)$ we use
\begin{equation}\label{eq:elementwise}
    F = \psi(I_t) \odot \phi(G_t),
\end{equation}
where $\odot$ is element-wise multiplication and we constrain $\phi(G_t)$ to have the same shape as $\psi(I_t)$. We observe from experiments that fusing $\psi(I_t)$ and $\phi(G_t)$ with $\odot$ performs better than channel-wise concatenation. Next, we define a cropping function to extract a 3$\times$3$\times$3 template kernel as,
\begin{equation}
    K=\mathcal{C}(F, \mu_t, (3,3,3)).
\end{equation}
where the kernel is centered at $\mu_t$ after any potential feature downscaling. To encode the global image context better, we also extract another larger size kernel $K_{g}=\mathcal{C}(F, \mu_t, (7,11,11))$. Here we limit its size in the $z$-direction to be $7$ since the size of $I_t$ during model training is only $(32,384,384)$.

Following the traditional cross-correlation operation \cite{Bertinetto2016SiameseFC}, we define the correspondence map as,
\begin{equation} \label{eq:cc}
    M = (K \star S) + (K_{g} \star S), 
\end{equation}
where $S=\psi(I_s)\odot\phi(G_s)$ and $+$ is the element-wise sum. Unfortunately, a direct use of $K_{g}$ introduces a heavy computational load. We propose to decompose $K_{g}$ along the axial, coronal, and sagittal directions and obtain flattened kernels as $K_{g,z}\in\mathbf{R}^{(1,11,11)}$, $K_{g,x}\in\mathbf{R}^{(7,1,11)}$, and $K_{g,y}\in\mathbf{R}^{(7,11,1)}$, where we omit the dimensions of batch size for clarity. As \Fig{vlt} demonstrates, the proposed \ac{my_speedup} layer performs the flattening using learned 3D convolutions  configured to produce an output of identical size as the kernel, except with one dimension flattened. The resulting faster version of \Eq{cc} is
\begin{equation} \label{eq:learnable}
    M = (K \star S) + \sum_{i\in{x,y,z}}K_{g,i} \star S.
\end{equation}
 We also tested kernel decomposition by simply extracting the middle ``slices'' of $K_{g}$ along the three dimensions, but it did not perform as well as the learned flattening operations.

Adding back the scale subscripts, the final output is a probability map:
\begin{equation}
    \hat{Y} = \sigma(W^T(M_1 + U_2 + U_3) + b),
\end{equation}
where $\sigma(\cdot)$ is the Sigmoid function, $W$ and $b$ are parameters of the final fully convolutional layer, $U_2$ is $M_2$ up-scaled by $(1,2,2)$, and $U_3$ is $M_3$ up-scaled by $(1,4,4)$. The predicted lesion center $\mu_p$ is the index of the global maximum in $\hat{Y}$.

\section{Supervised and Self-Supervised Learning}
\ac{my_method} is capable of both supervised and \acf{SSL}. It is flexible enough to learn from paired annotations, when enough are available, and to also use efficient self-supervised learning. 

\subsection{Supervised Learning}
\label{sec:supervised}
Based on the introduced network architecture, $\hat{Y}$, the output of \ac{my_method} is a dense probability map representing the likelihood of each location to be the target lesion center. Therefore, we define the ground truth as a Gaussian kernel centered at the target location $\mu_s$. Formally, we first define $Y=\mathcal{G}(\mu_s,r_s)$ and then downsize it to match the dimensions of $\hat{Y}$. We use focal loss \cite{Lin2020FocalLoss,Zhou2019CenterNet} in training:
\begin{equation} \label{eq:centerLoss}
    \mathcal{L}_{sl} = \sum_{i}
    \begin{cases}
        (1 - \hat{y}_{i})^{\alpha} 
        \log(\hat{y}_{i}) & \!\text{if}\ y_{i}=1 \\
        (1-y_{i})^{\beta} 
        (\hat{y}_{i})^{\alpha}   
        \log(1-\hat{y}_{i})
        & \!\text{otherwise}
    \end{cases} \textrm{,}
\end{equation}
where $y_i$ and $\hat{y}_i$ are the $i$-th voxels in $Y$ and $\hat{Y}$, respectively, and $\alpha=2$ and $\beta=4$ are focal-loss hyper-parameters~\cite{Lin2020FocalLoss,Zhou2019CenterNet}. The ground-truth heat map is $<1$ everywhere except at the lesion center voxel. So that the training can converge quickly, it ignores hard voxels that are near $\mu_s$. 

{\bf Center augmentation.} In practice, labels from clinicians may not represent the exact lesion centers. The provided location, $\mu_t$, may shift inside the central area. Therefore, to increase model robustness we train \ac{my_method} with random location shifts. This is achieved by adding $\mu_t$ with $\Delta\mu_t$, which is randomly sampled from the sphere $\|\Delta\mu_t\|_2\leq0.25r_t$.

\subsection{Self-Supervised Learning}
Since our proposed \ac{my_method} is built upon Siamese pair-wise comparison, it inherently supports learning with self-supervision. The key insight is that effective visual representation for object recognition can be learned by comparing the template image, $I_t$, with its augmented counterparts. With $I_t$, we implement data augmentations including (1) elastic deformations at  random scales ranging from 0 to 0.25, (2) rotations in the $xy$-plane with a random angle ranging from -10 to 10 degrees, (3) random scales ranging from 0.75 to 1.25, (4) random crops, (5) add Gaussian noise with zero mean and a random variance ranging from 0 to 0.05, and (6) Gaussian blurring with a random sigma ranging from 0.5 to 1.5 \cite{Isensee2018nnUNet}. Each augmentation individually takes place with the probability of 0.5. For clarity, we define $\mathcal{T}_{\text{aug}}$ as any combination of the data augmentations. Therefore, each self-supervised image ``pair'' comprises $I_t$ and $\mathcal{T}_{\text{aug}}(I_t)$ with corresponding anatomical signals of $G_t$ and $\mathcal{T}_{\text{aug}}(G_t)$. The same training procedure as supervised learning can then be followed. It is worth mentioning that our \ac{SSL} strategy shares a similar spirit with recent contrastive learning studies that matches an image with its transformed version \cite{Chen2020SimCLR}, but in the pixel-level.

We select non-longitudinal images from DeepLesion~\cite{Yan2018DeepLesion} and use the bounding box annotations as $\mu_t$ and $r_t$. When bounding box annotations are not available, the template lesions can be extracted by applying a pre-trained universal lesion detector on $I_t$ and randomly selecting top-scoring proposals. However, we do not explore that here. 

Limited by GPU memory, when combining the supervised learning with \ac{SSL}, we switch the training of \ac{my_method} between both schemes as:
\begin{equation}
    \mathcal{L}_{mix} = 
    \begin{cases}
        \mathcal{L}_{ssl} & \!\text{if}\ \lambda\leq \tau \\
        \mathcal{L}_{sl} & \!\text{otherwise}
    \end{cases} \textrm{,} \label{eq:mix}
\end{equation}
where $\lambda\in[0,1]$ is a random number and we empirically set the threshold $\tau$ to 0.25 in our experiments.

\section{Experiments}
\subsection{Datasets} \label{sec:datasets}
{\bf DeepLesion} is a large-scale \ac{CT} database of lesions released by the National Institute of Health (NIH) in 2018 \cite{Yan2018DeepLesion,Yan2018LesionGraph}. It contains over 30 thousand lesions and each lesion is associated with a size measurement defined by the \ac{RECIST} \cite{Eisenhauer2009RECIST}. The \ac{RECIST} measurement consists of two diameters: the longest diameter followed by the longest diameter that is perpendicular to the first one. Both diameters are drawn by doctor in a manually selected axial slice. Based on this measurement, we define the ground truth lesion center $\mu$ to be the mean of diameters' four end points and the radius $r$ is approximated to be the half of the longest diameters. In total, our publicly released \ac{my_dataset} dataset inherits 3008, 403, and 480 lesion pairs from the DeepLesion's train, validate, and test splits, respectively. 

From Chang Gung Memorial Hospital (IRB 202000584A3C601), we also collected an external validation set that consists of 536 lesions from 100 longitudinal studies of 86 patients, including 45 cervical, 27 endometrial, and 14 ovarian cancers (mean age, 53.3 years). The median time interval between CT studies was 217 days (range, 8-2304 days). We apply the best \ac{my_method} configuration, developed on the DeepLesion dataset, to track the corresponding target lesions, if they exist. To assess the tracking accuracy we measured the acceptance rate of an board-certificated radiologist with over 10 years of clinical practice experience.

\subsection{Evaluation Metrics}
For an annotated pair of lesion $a$ and $b$, we evaluate tracking both from $a$ to $b$ and from $b$ to $a$. Therefore, in total, we have 906 and 960 directed lesion pairs in the validation and test sets, respectively. We define a center point matching (CPM) accuracy, which represents the percentage of correctly matched lesions. A match will be counted correct when the Euclidean distance between the ground truth center and the predicted center is smaller than a threshold. We first set the threshold to be the corresponding lesion radius and refer the matching accuracy CPM@Radius or simply CPM. However this threshold is not tight enough to differentiate trackers as some lesions have large sizes. We then use an adaptive threshold $min(r,10mm)$ to limit the allowed maximum offset in large lesions and we refer to this matching accuracy as CPM@10$mm$. We empirically use 10$mm$ because 55\% lesions in the test set have larger than 10$mm$ radiuses.

We also measure the absolute offset between ground truth and predicted centers in $mm$ and report the mean Euclidean distance (MED) and its projections MED$_X$, MED$_Y$, MED$_Z$ in each direction. The speed of trackers is counted using seconds per volume (spv). 

\subsection{Comparisons with State-of-the-art Approaches} 

\begin{figure*}[t!]
    \centering
    \includegraphics[width=0.98\linewidth]{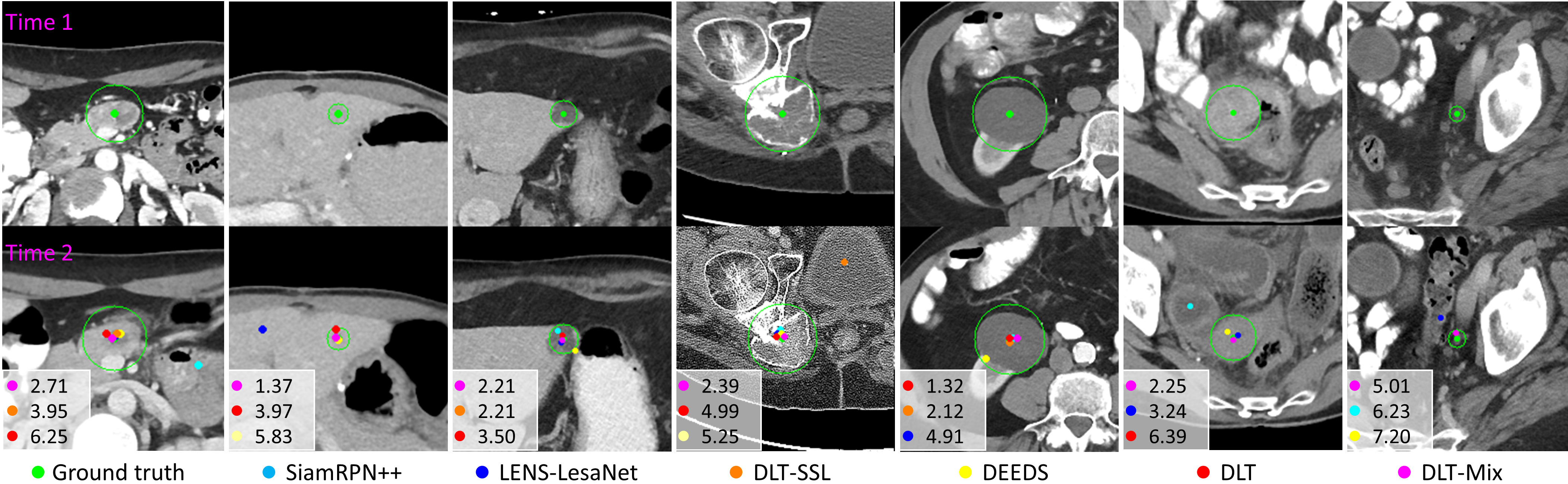}
    \caption{Comparisons of our methods with three state-of-the-art trackers. The top 3 closest to center distances are reported in $mm$.}
    \label{fig:examples_more}
\end{figure*}

\begin{table*}[t!]
  \small
  \centering
  \begin{tabular}{lccccccc}
   \hline
   Method & CPM@ & CPM@ & MED$_X$ & MED$_Y$ & MED$_Z$ & MED & speed \\ 
   & 10$mm$ & Radius & ($mm$) & ($mm$) & ($mm$) & ($mm$) & (spv) \\
   \hline
   Affine \cite{Marstal2016SimpleElastix}                        & 48.33 & 65.21 & 4.1$\pm$5.0 & 5.4$\pm$5.6 & 7.1$\pm$8.3 & 11.2$\pm$9.9 & 1.82 \\
   VoxelMorph \cite{Balakrishnan2018Reg}                         & 49.90 & 65.59 & 4.6$\pm$6.7 & 5.2$\pm$7.9 & 6.6$\pm$6.2 & 10.9$\pm$10.9 & {\bf 0.46} \\
   LENS-LesionGraph \cite{Yan2020Lens,Yan2018LesionGraph}        & 63.85 & 80.42 & {\bf 2.6$\pm$4.6} & 2.7$\pm$4.5 & 6.0$\pm$8.6 & 8.0$\pm$10.1 & 4.68 \\
   VULD-LesionGraph \cite{Cai2020VULD,Yan2018LesionGraph}        & 64.69 & 76.56 & 3.5$\pm$5.2 & 4.1$\pm$5.8 & 6.1$\pm$8.8 & 9.3$\pm$10.9 & 9.07 \\
   VULD-LesaNet \cite{Cai2020VULD,Yan2019LesaNet}                & 65.00 & 77.81 & 3.5$\pm$5.3 & 4.0$\pm$5.7 & 6.0$\pm$8.7 & 9.1$\pm$10.8 & 9.05 \\
   SiamRPN$++$ \cite{Li2019SiamRPN++}                            & 68.85 & 80.31 & 3.8$\pm$4.8 & 3.8$\pm$4.8 & 4.8$\pm$7.5 & 8.3$\pm$9.2 & 2.24 \\
   LENS-LesaNet \cite{Yan2020Lens,Yan2019LesaNet}                & 70.00 & 84.58 & 2.7$\pm$4.8 & {\bf 2.6$\pm$4.7} & 5.7$\pm$8.6 & 7.8$\pm$10.3 & 4.66 \\
   \ac{my_method}-SSL                                            & 71.04 & 81.52 & 3.8$\pm$5.3 & 3.7$\pm$5.5 & 5.4$\pm$8.4 & 8.8$\pm$10.5 & 3.57 \\
   DEEDS \cite{Heinrich2013DeedsBCV}                             & 71.88 & 85.52 & 2.8$\pm$3.7 & 3.1$\pm$4.1 & 5.0$\pm$6.8 & 7.4$\pm$8.1 & 15.3 \\
   \ac{my_method}-Mix                                            & 78.65 & {\bf 88.75} & 3.1$\pm$4.4 & 3.1$\pm$4.5 & 4.2$\pm$7.6 & 7.1$\pm$9.2 & 3.54 \\
   \ac{my_method}                                                & {\bf 78.85} & 86.88 & 3.5$\pm$5.6 & 2.9$\pm$4.9 & {\bf 4.0$\pm$6.1} & {\bf 7.0$\pm$8.9} & 3.58 \\
   \hline
  \end{tabular}
  \caption{Comparisons between the proposed \ac{my_method} and state-of-the-art approaches.}
  \label{tbl:cpm}
\end{table*}

\noindent {\bf Traditional registration approaches.} We use both the widely used rigid affine registration method \cite{Marstal2016SimpleElastix} and DEEDS \cite{Heinrich2013DeedsBCV} deformable registration. The latter is considered the state-of-the-art deformable approach for CT registration \cite{Xu2016SixRegistration}. The implementation that we use is optimized in C++ \cite{Heinrich2015DeedsBCV} and the \ac{CT} volumes have been resampled to the isotropic resolution of 2$mm$.

\noindent {\bf Learning based registration.} We use VoxelMorph \cite{Balakrishnan2019VoxelMorph}, which is a general deep learning framework for deformable medical image registration that can deliver state-of-the-art performance with a much faster speeds than traditional approaches. We train VoxelMorph with image pairs from \ac{my_dataset}. Image pairs are first aligned by affine registration and then resampled to 0.8$mm$ by 0.8$mm$ in $xy$-plane with a slice thickness of 2$mm$. The same image resolution is applied to all of the following experiments.

\noindent {\bf Tracking by detector learning.} These approaches first detect lesion candidates. Then, an image encoder is used to project both the template lesion and the detected candidates into feature vectors. Lastly, a nearest neighbor classifier is applied to identify the tracked lesion. We tested the detector with the 2D LENS \cite{Yan2018DeepLesion} and 3D VULD \cite{Cai2020VULD} detectors, both of which report good performance on DeepLesion. As for the image encoder, we tested LesionGraph \cite{Yan2018LesionGraph} and LesaNet \cite{Yan2019LesaNet}, which are also developed from DeepLesion for lesion attribute description. Therefore, we evaluate four baselines, \ie{}, LENS-LesionGraph, LENS-LesaNet, VULD-LesionGraph, and VULD-LesaNet. 

\noindent {\bf Tracking by similarity learning.} We adapt SiamRPN++ \cite{Li2019SiamRPN++} with 3D DenseFPN so that it can process \ac{CT} images and perform fair comparison with \ac{my_method}. The largest size of the template kernel is $(3,5,5)$ for computational efficiency. 

\noindent {\bf \ac{my_method} and its variants.} \ac{my_method} is trained using \ac{my_dataset}. \ac{my_method}-SSL is trained using only \ac{SSL} with non-longitudinal training images of DeepLesion that do not exist in \ac{my_dataset}. \ac{my_method}-Mix is trained with a combination of supervised and self-supervised learning, which is defined by \Eq{mix}.

\begin{table}[t!]
   \small
   \centering 
   \begin{tabular}{lcc}
      \hline 
      Method & CPM@10$mm$ & MED ($mm$) \\
      \hline
      SiamRPN++ \cite{Li2019SiamRPN++}       & 51.27 ($\downarrow$ 17.6) & 10.6$\pm$10.3 ($\uparrow$ 2.3) \\
      DEEDS \cite{Heinrich2013DeedsBCV}      & 53.85 ($\downarrow$ 18.0) & 9.8$\pm$8.9 ($\uparrow$ 2.4) \\
      \ac{my_method}-SSL                     & 64.24 ($\downarrow$ 6.80) & 10.0$\pm$11.4 ($\uparrow$ 1.2) \\
      \ac{my_method}                         & 70.36 ($\downarrow$ 8.49) & 8.1$\pm$8.7 ($\uparrow$ 1.2) \\
      \ac{my_method}-Mix                     & {\bf 75.03 ($\downarrow$ 3.62)} & {\bf 8.0$\pm$10.5 ($\uparrow$ 0.9)} \\
      \hline
   \end{tabular}
   \caption{Robustness evaluation. $\downarrow$ and $\uparrow$ demonstrate decrease and increase of measurements, respectively, compared with the values reported in \Table{cpm}.}
   \label{tbl:robustness}
\end{table}

\noindent {\bf Results.} \Table{cpm} shows the comparative results. With CPM@10$mm$, \ac{my_method} and \ac{my_method}-Mix achieve the first and second places, respectively, leading DEEDS at the third place by over 6\%. \ac{my_method}-SSL is at the 4th place outperforming its \ac{SSL} counterparts, \ie{}, affine registration and VoxelMorph, by over 20\%. With CPM@Radius, \ac{my_method}-Mix is the best tracker, and it outperforms DEEDs and SiamRPN++ by 3.2\% and 8.4\%, respectively. With MED, \ac{my_method} performs the best. We notice that LENS-LesionGraph outperforms \ac{my_method} in MED$_X$ by 0.9$mm$ because LENS is a 2D lesion detector with a bounding-box regression layer, which is dedicated to locating the lesion accurately in the $xy$-plane. Similarly, LENS-LesaNet outperforms \ac{my_method} by 0.3$mm$ in MED$_Y$. However, in MED$_Z$, \ac{my_method} greatly outperforms LENS-LesionGraph and LENS-LesaNet by 2$mm$ and 1.7$mm$, respectively, showing the importance of 3D DenseFPN. In terms of speed, affine registration and VoxelMorph are the top 2 methods but they are not as accurate as the others. Among the top 3 methods, \ac{my_method} and \ac{my_method}-Mix run about 4 times faster than DEEDS on the DeepLesion dataset. \Fig{examples_more} shows seven visual examples of lesion tracking, where the results produced by our trackers are closer to the ground truth than others.

\noindent {\bf Robustness evaluation.} In this experiment, we simulate human inputs. In testing, we shift the template center $\mu_t$ with $\Delta\mu_t$, which is randomly sampled from the sphere $\|\Delta\mu\|_2\leq$0.25$r_t$. For each directed lesion pair, 9 shifted centers together with the original center are stored. In total, we create 9060 and 9600 directed lesion pairs from the validation and test sets, respectively. With these augmented lesion pairs, we evaluate trackers to see if they are robust with inaccurate human inputs or not.  

\Table{robustness} shows the results. \ac{my_method}-Mix is in the first place for both CPM and MED metrics. DEEDS turns out to be the most vulnerable method with 18\% drop in CPM and 2.4$mm$ increase in MED. In comparison, \ac{my_method}-Mix only drops 3.62\% in CPM and increases only 0.9$mm$ in MED. Additionally, \ac{my_method}-SSL is more robust than \ac{my_method} in CPM, demonstrating the benefit of \ac{SSL} in training robust trackers.

\begin{table}[t!]
    \small
    \centering
    \begin{tabular}{lccccccc}
    \hline
    & \multicolumn{2}{c}{\Eq{learnable}: $K_{g}$} & $\psi, \phi$ & \Eq{ase_heatmap}: $G$ & test & speed \\
    id    & size & learn        & dim.& size ($n$) & MED & spv \\
    \hline
    $a$ &  N/A	   & N/A        & 64 & 4 & 9.3 & 1.44 \\
    $b$ &  7,7,7   & \checkmark & 64 & 4 & 9.4 & 2.38 \\
    $c$ &  7,15,15 & \checkmark & 64 & 4 & 7.7 & 24.1 \\
    $d$ &  7,11,11 & \checkmark & 64 & 2 & 7.4 & 3.51 \\
    $e$ &  7,11,11 & \checkmark & 64 & 8 & 8.5 & 3.51 \\
    $f$ &  7,11,11 & \checkmark & 32 & 4 & 8.7 & 2.25 \\
    $g$ &  7,11,11 & \checkmark &128 & 4 & 7.9 & 5.83 \\
    $h$ &  7,11,11 & \checkmark & 64 & N/A & 9.3 & 3.51 \\ 
    $i$ &  7,11,11 & \XSolidBrush & 64 & 4 & 9.3 & 3.51 \\
    $j$ &  7,11,11 & \checkmark & 64 & 4 & 7.9 & 3.51 \\
    \hline
    \end{tabular}
    \caption{Parameter analysis of the proposed components.}
    \label{tbl:pa}
\end{table}

\noindent {\bf Parameter Analysis.} \Table{pa} presents our parameter analysis for different model configurations, with model $j$ representing our final configuration without the multiplication fusion of \Eq{elementwise} or the center augmentation of \Sec{supervised}. We present test results, but note that our model selection was based off of our validation (which can be found in the supplementary material). Model $a$ is identical to our final model, except that the global kernel has been disabled, resulting in significant MED increases and demonstrating the importance of the global kernel. Models $b$ and $c$ explore different global kernel sizes, indicating performance can vary somewhat, but is not overly sensitive to the choice. However, too large of a kernel results in an order of magnitude greater runtime, justifying our choice of a $(7,11,11)$ kernel. As model $e$ demonstrates,  when the \ac{anatomy_net} heat map of \Eq{ase_heatmap} covers too large of an area it can lose its specificity, resulting in performance degradation. Models $f$ and $g$ show the effect of different embedding feature dimensions, again showing that performance is not overly sensitive to this choice, as long as the embedding dimension is large enough. In terms of the need for the anatomy signal of \ac{anatomy_net}, model $h$ demonstrates its removal considerably increases the MED. Finally, model $i$'s performance shows that the learnable decomposition of \Eq{learnable} is critical for accurate tracking. Adding \Eq{elementwise} and center augmentation to model $j$ results in our final configuration featured in \Table{cpm}.

\subsection{Impact on Downstream Measurements} \label{sec:measure}
In this experiment, we compare trackers with downstream size measurements. We use a pre-trained model, OneClick \cite{Tang2020OneClick} that takes the image $I_s$ and the predicted lesion center $\mu_p$ as its inputs and regresses the \ac{RECIST} diameters of the target lesion. For simplicity, we only compare the long diameters. We use evaluation metrics including mean absolute error (MAE) in $mm$, growth accuracy, and treatment response accuracy. With the template diameter $d_t$, search diameter $d_s$, and OneClick predicted diameter $d_p$, we define $d_p$ as a correct growth prediction, if and only if the inequality ($d_s$-$d_t$)($d_p$-$d_t$)$>$0 holds. The growth accuracy represents the percentage of correct growth predictions. The treatment response, $\rho$=($d_s$-$d_t$)/$d_t$, is defined based on the \ac{RECIST} guideline \cite{Eisenhauer2009RECIST}, which classifies a treatment response as partial response if $\rho$ $\leq$-0.3, as progressive disease if $\rho$ $\geq$0.2, or as stable disease if $\rho\in$(-0.3,0.2). We then predict treatment response using $\rho_p$=($d_p$-$d_t$)/$d_t$.

We tested \ac{my_method}, DEEDS, and manual inputs, \ie{} the ground truth lesion centers. \Table{one_click} shows the results. \ac{my_method} outperforms DEEDS in MAE by 0.22$mm$, which is an 8\% improvement. Compared with manual inputs, \ac{my_method} exhibits the same growth accuracy and is only 0.46\% lower in the treatment response accuracy.

\begin{table}[t!]
   \small
   \centering 
   \begin{tabular}{lccc} 
      \hline 
      Input & MAE & Growth & Response \\
      generator & ($mm$) & acc. (\%) & acc. (\%) \\
      \hline
      DEEDS \cite{Heinrich2013DeedsBCV} & 2.69$\pm$4.12 & 78.02 & 84.17 \\
      \ac{my_method} & 2.47$\pm$3.58 & {\bf 79.69} & 85.10 \\
      Manual inputs   & {\bf 2.31$\pm$3.16} & {\bf 79.69} & {\bf 85.56} \\
      \hline
   \end{tabular}
   \caption{Impact on automatic lesion size measurement when using the OneClick \cite{Tang2020OneClick} model.}
   \label{tbl:one_click}
\end{table}

\begin{table}[t!]
   \small
   \centering 
   \begin{tabular}{lcc}
      \hline 
      Method & CPM@Radius & speed (spv) \\
      \hline 
      DEEDS \cite{Heinrich2013DeedsBCV} & 85.6 & 67.1$\pm$17.8 \\
      \ac{my_method} & {\bf 88.4} & {\bf 4.7$\pm$0.35} \\
      \hline
   \end{tabular}
   \caption{External evaluation.}
   \label{tbl:external}
\end{table}

\noindent {\bf External Evaluation.} We further invite a board-certified radiologist to manually assess \ac{my_method} with 100 longitudinal studies recruited from real-life clinical workflows. The user provides binarized responses, \ie{}, \textit{inside-} or \textit{outside-lesion} for the CPM@Radius metric. We compared the tracking results of \ac{my_method} with DEEDS in \Table{external}. \ac{my_method} delivers 88.4\% CPM accuracy and outperforms DEEDS by 2.8\%. Besides, as a more \textit{true-to-life} measurement, \ac{my_method} requires only 4.67 seconds to process a whole body \ac{CT}, which is over 14 times faster than DEEDS. These results also underscore the value of our \ac{my_dataset} dataset.

\section{Conclusion \& Discussion}
In this work, we introduce a new public benchmark for lesion tracking and present \ac{my_method} as our solution. Due to the different setup of medical applications, \ac{my_method} differs from general visual trackers in two aspects. First, \ac{my_method} does not regress bounding boxes for target lesions because as mentioned in \Sec{measure}, the lesion size can be accurately predicted by the down stream measurement module. Second, \ac{my_method} does not perform long-term tracking because time points in longitudinal studies is much less than general videos. Also, manual calibration occurs much more often in lesion tracking than general object tracking.

Our presented \ac{my_method} has been demonstrated effective for lesion tracking, outperforming a comprehensive set of baselines that represent various tracking strategies. \ac{my_method} can be trained via either supervised or self-supervised learning, where the combination of both training schemes results in the best performance and robustness. We benchmark the task of lesion tracking on our \ac{my_dataset} dataset which will be made available upon request.  

{\small
\bibliographystyle{ieee_fullname}
\bibliography{egbib}

\begin{thebibliography}{10}\itemsep=-1pt

\bibitem{Ardila2019LungCancer}
Diego Ardila, Atilla~P Kiraly, Sujeeth Bharadwaj, Bokyung Choi, Joshua~J
  Reicher, Lily Peng, Daniel Tse, Mozziyar Etemadi, Wenxing Ye, Greg Corrado,
  et~al.
\newblock End-to-end lung cancer screening with three-dimensional deep learning
  on low-dose chest computed tomography.
\newblock {\em Nature medicine}, 25(6):954--961, 2019.

\bibitem{Avidan2004SVT}
Shai Avidan.
\newblock Support vector tracking.
\newblock {\em IEEE Trans. Pattern Anal. Mach. Intell.}, 26(8):1064--1072,
  2004.

\bibitem{Babenko2011TrackingMIL}
Boris Babenko, Ming{-}Hsuan Yang, and Serge~J. Belongie.
\newblock Robust object tracking with online multiple instance learning.
\newblock {\em IEEE Trans. Pattern Anal. Mach. Intell.}, 33(8):1619--1632,
  2011.

\bibitem{Balakrishnan2018Reg}
Guha Balakrishnan, Amy Zhao, Mert Sabuncu, John Guttag, and Adrian~V. Dalca.
\newblock An unsupervised learning model for deformable medical image
  registration.
\newblock {\em IEEE Conf. Comput. Vis. Pattern Recog.}, pages 9252--9260, 2018.

\bibitem{Balakrishnan2019VoxelMorph}
Guha Balakrishnan, Amy Zhao, Mert~R. Sabuncu, John~V. Guttag, and Adrian~V.
  Dalca.
\newblock Voxelmorph: {A} learning framework for deformable medical image
  registration.
\newblock {\em IEEE Trans. Med. Imaging}, 38(8):1788--1800, 2019.

\bibitem{Bertinetto2016STAPLE}
Luca Bertinetto, Jack Valmadre, Stuart Golodetz, Ondrej Miksik, and Philip
  H.~S. Torr.
\newblock Staple: Complementary learners for real-time tracking.
\newblock In {\em IEEE Conf. Comput. Vis. Pattern Recog.}, pages 1401--1409.
  2016.

\bibitem{Bertinetto2016SiameseFC}
Luca Bertinetto, Jack Valmadre, Jo{\~{a}}o~F. Henriques, Andrea Vedaldi, and
  Philip H.~S. Torr.
\newblock Fully-convolutional siamese networks for object tracking.
\newblock In {\em Eur. Conf. Comput. Vis. Worksh.}, pages 850--865, 2016.

\bibitem{Bolme2010MOSSE}
David~S. Bolme, J.~Ross Beveridge, Bruce~A. Draper, and Yui~Man Lui.
\newblock Visual object tracking using adaptive correlation filters.
\newblock In {\em IEEE Conf. Comput. Vis. Pattern Recog.}, pages 2544--2550.
  2010.

\bibitem{Bromley1993Siamese}
Jane Bromley, James~W. Bentz, L{\'{e}}on Bottou, Isabelle Guyon, Yann LeCun,
  Cliff Moore, Eduard S{\"{a}}ckinger, and Roopak Shah.
\newblock Signature verification using {A} "siamese" time delay neural network.
\newblock {\em Int. J. Pattern Recognit. Artif. Intell.}, 7(4):669--688, 1993.

\bibitem{Cai2020LesionHarvester}
J. {Cai}, A.~P. {Harrison}, Y. {Zheng}, K. {Yan}, Y. {Huo}, J. {Xiao}, L.
  {Yang}, and L. {Lu}.
\newblock Lesion-harvester: Iteratively mining unlabeled lesions and
  hard-negative examples at scale.
\newblock {\em IEEE Trans. Med. Imaging}, pages 1--1, 2020.

\bibitem{Cai2018SlicePropagation}
Jinzheng Cai, Youbao Tang, Le Lu, Adam~P. Harrison, Ke Yan, Jing Xiao, Lin
  Yang, and Ronald~M. Summers.
\newblock Accurate weakly-supervised deep lesion segmentation using large-scale
  clinical annotations: Slice-propagated 3d mask generation from 2d {RECIST}.
\newblock In {\em Medical Image Computing and Computer Assisted Intervention},
  pages 396--404. 2018.

\bibitem{Cai2020VULD}
Jinzheng Cai, Ke Yan, Chi{-}Tung Cheng, Jing Xiao, Chien{-}Hung Liao, Le Lu,
  and Adam~P. Harrison.
\newblock Deep volumetric universal lesion detection using light-weight pseudo
  3d convolution and surface point regression.
\newblock In {\em Medical Image Computing and Computer Assisted Intervention},
  pages 3--13. 2020.

\bibitem{Carreira2017I3D}
Jo{\~{a}}o Carreira and Andrew Zisserman.
\newblock Quo vadis, action recognition? {A} new model and the kinetics
  dataset.
\newblock In {\em IEEE Conf. Comput. Vis. Pattern Recog.}, pages 4724--4733.
  2017.

\bibitem{Gomariz2019LiverLandmark}
Alvaro~Gomariz Carrillo, Weiye Li, Ece Ozkan, Christine Tanner, and Orcun
  Goksel.
\newblock Siamese networks with location prior for landmark tracking in liver
  ultrasound sequences.
\newblock In {\em IEEE Int. Symposium on Biomedical Imaging}, pages 1757--1760.
  2019.

\bibitem{Chen2018DeepLab}
Liang{-}Chieh Chen, George Papandreou, Iasonas Kokkinos, Kevin Murphy, and
  Alan~L. Yuille.
\newblock Deeplab: Semantic image segmentation with deep convolutional nets,
  atrous convolution, and fully connected crfs.
\newblock {\em IEEE Trans. Pattern Anal. Mach. Intell.}, 40(4):834--848, 2018.

\bibitem{Chen2020SimCLR}
Ting Chen, Simon Kornblith, Mohammad Norouzi, and Geoffrey~E. Hinton.
\newblock A simple framework for contrastive learning of visual
  representations.
\newblock {\em CoRR}, abs/2002.05709, 2020.

\bibitem{Cicek2016UNet3D}
{\"{O}}zg{\"{u}}n {\c{C}}i{\c{c}}ek, Ahmed Abdulkadir, Soeren~S. Lienkamp,
  Thomas Brox, and Olaf Ronneberger.
\newblock 3d u-net: Learning dense volumetric segmentation from sparse
  annotation.
\newblock In {\em Medical Image Computing and Computer Assisted Intervention},
  pages 424--432, 2016.

\bibitem{Danelljan2015DeepSRDCF}
Martin Danelljan, Gustav H{\"{a}}ger, Fahad~Shahbaz Khan, and Michael Felsberg.
\newblock Convolutional features for correlation filter based visual tracking.
\newblock In {\em Int. Conf. Comput. Vis. Worksh.}, pages 621--629. 2015.

\bibitem{Danelljan2014DSST}
Martin Danelljan, Fahad~Shahbaz Khan, Michael Felsberg, and Joost van~de
  Weijer.
\newblock Adaptive color attributes for real-time visual tracking.
\newblock In {\em IEEE Conf. Comput. Vis. Pattern Recog.}, pages 1090--1097.
  2014.

\bibitem{Eisenhauer2009RECIST}
E. Eisenhauer, P. Therasse, J. Bogaerts, and et al.
\newblock New response evaluation criteria in solid tumours: revised recist
  guideline (version 1.1).
\newblock {\em European journal of cancer}, 45(2):228--247, 2009.

\bibitem{Girshick2014RCNN}
Ross~B. Girshick, Jeff Donahue, Trevor Darrell, and Jitendra Malik.
\newblock Rich feature hierarchies for accurate object detection and semantic
  segmentation.
\newblock In {\em IEEE Conf. Comput. Vis. Pattern Recog.}, pages 580--587.
  2014.

\bibitem{Heinrich2013DeedsBCV}
M.~P. {Heinrich}, M. {Jenkinson}, M. {Brady}, and J.~A. {Schnabel}.
\newblock Mrf-based deformable registration and ventilation estimation of lung
  ct.
\newblock {\em IEEE Trans. Med. Imaging}, 32(7):1239--1248, 2013.

\bibitem{Heinrich2015DeedsBCV}
Mattias~P. Heinrich, Oskar Maier, and Heinz Handels.
\newblock Multi-modal multi-atlas segmentation using discrete optimisation and
  self-similarities.
\newblock In {\em IEEE Int. Symposium on Biomedical Imaging}, pages 27--30.
  2015.

\bibitem{Henriques2012Circulant}
Jo{\~{a}}o~F. Henriques, Rui Caseiro, Pedro Martins, and Jorge~P. Batista.
\newblock Exploiting the circulant structure of tracking-by-detection with
  kernels.
\newblock In {\em Eur. Conf. Comput. Vis.}, pages 702--715. 2012.

\bibitem{Huang2017DenseNet}
Gao Huang, Zhuang Liu, Laurens van~der Maaten, and Kilian~Q. Weinberger.
\newblock Densely connected convolutional networks.
\newblock In {\em IEEE Conf. Comput. Vis. Pattern Recog.}, pages 2261--2269.
  2017.

\bibitem{Isensee2018nnUNet}
Fabian Isensee, Jens Petersen, Andr{\'{e}} Klein, David Zimmerer, Paul~F.
  Jaeger, Simon Kohl, Jakob Wasserthal, Gregor Koehler, Tobias Norajitra,
  Sebastian~J. Wirkert, and Klaus~H. Maier{-}Hein.
\newblock nnu-net: Self-adapting framework for u-net-based medical image
  segmentation.
\newblock {\em CoRR}, abs/1809.10486, 2018.

\bibitem{Jiang2020ElixirNet}
Chenhan Jiang, Shaoju Wang, Xiaodan Liang, Hang Xu, and Nong Xiao.
\newblock Elixirnet: Relation-aware network architecture adaptation for medical
  lesion detection.
\newblock In {\em AAAI}, pages 11093--11100. 2020.

\bibitem{Krizhevsky2012AlexNet}
Alex Krizhevsky, Ilya Sutskever, and Geoffrey~E. Hinton.
\newblock Imagenet classification with deep convolutional neural networks.
\newblock In {\em Adv. Neural Inform. Process. Syst.}, pages 1106--1114, 2012.

\bibitem{Li2019SiamRPN++}
Bo Li, Wei Wu, Qiang Wang, Fangyi Zhang, Junliang Xing, and Junjie Yan.
\newblock Siamrpn++: Evolution of siamese visual tracking with very deep
  networks.
\newblock In {\em {IEEE} Conference on Computer Vision and Pattern Recognition,
  {CVPR} 2019, Long Beach, CA, USA, June 16-20, 2019}, pages 4282--4291. 2019.

\bibitem{Lin2017FPN}
Tsung{-}Yi Lin, Piotr Doll{\'{a}}r, Ross~B. Girshick, Kaiming He, Bharath
  Hariharan, and Serge~J. Belongie.
\newblock Feature pyramid networks for object detection.
\newblock In {\em IEEE Conf. Comput. Vis. Pattern Recog.}, pages 936--944.
  2017.

\bibitem{Lin2020FocalLoss}
Tsung{-}Yi Lin, Priya Goyal, Ross~B. Girshick, Kaiming He, and Piotr
  Doll{\'{a}}r.
\newblock Focal loss for dense object detection.
\newblock {\em {IEEE} Trans. Pattern Anal. Mach. Intell.}, 42(2):318--327,
  2020.

\bibitem{Liu2020COSD-CNN}
Fei Liu, Dan Liu, Jie Tian, Xiaoyan Xie, Xin Yang, and Kun Wang.
\newblock Cascaded one-shot deformable convolutional neural networks:
  Developing a deep learning model for respiratory motion estimation in
  ultrasound sequences.
\newblock {\em Med. Image Anal.}, 65:101793, 2020.

\bibitem{Liu2015PBV}
Ting Liu, Gang Wang, and Qingxiong Yang.
\newblock Real-time part-based visual tracking via adaptive correlation
  filters.
\newblock In {\em IEEE Conf. Comput. Vis. Pattern Recog.}, pages 4902--4912.
  2015.

\bibitem{Long2015FCN}
Jonathan Long, Evan Shelhamer, and Trevor Darrell.
\newblock Fully convolutional networks for semantic segmentation.
\newblock In {\em IEEE Conf. Comput. Vis. Pattern Recog.}, pages 3431--3440.
  2015.

\bibitem{Ma2015CF2}
Chao Ma, Jia{-}Bin Huang, Xiaokang Yang, and Ming{-}Hsuan Yang.
\newblock Hierarchical convolutional features for visual tracking.
\newblock In {\em Int. Conf. Comput. Vis.}, pages 3074--3082. 2015.

\bibitem{Marstal2016SimpleElastix}
Kasper Marstal, Floris~F. Berendsen, Marius Staring, and Stefan Klein.
\newblock Simpleelastix: {A} user-friendly, multi-lingual library for medical
  image registration.
\newblock In {\em IEEE Conf. Comput. Vis. Pattern Recog. Worksh.}, pages
  574--582. 2016.

\bibitem{Miao2016Registration}
Shun Miao, Z.~Jane Wang, and Rui Liao.
\newblock A {CNN} regression approach for real-time 2d/3d registration.
\newblock {\em IEEE Trans. Med. Imaging}, 35(5):1352--1363, 2016.

\bibitem{Milletari2016VNet}
Fausto Milletari, Nassir Navab, and Seyed{-}Ahmad Ahmadi.
\newblock V-net: Fully convolutional neural networks for volumetric medical
  image segmentation.
\newblock In {\em 3DV}, pages 565--571. 2016.

\bibitem{Nam2016MDNet}
Hyeonseob Nam and Bohyung Han.
\newblock Learning multi-domain convolutional neural networks for visual
  tracking.
\newblock In {\em IEEE Conf. Comput. Vis. Pattern Recog.}, pages 4293--4302.
  2016.

\bibitem{RafaelPalou2020Noudle}
Xavier Rafael-Palou, Anton Aubanell, Ilaria Bonavita, Mario Ceresa, Gemma
  Piella, Vicent Ribas, and Miguel~A. {González Ballester}.
\newblock Re-identification and growth detection of pulmonary nodules without
  image registration using 3d siamese neural networks.
\newblock {\em Med. Image Anal.}, 67:101823, 2021.

\bibitem{Raju2020Liver}
Ashwin Raju, Chi{-}Tung Cheng, Yuankai Huo, Jinzheng Cai, Junzhou Huang, Jing
  Xiao, Le Lu, Chien{-}Hung Liao, and Adam~P. Harrison.
\newblock Co-heterogeneous and adaptive segmentation from multi-source and
  multi-phase {CT} imaging data: {A} study on pathological liver and lesion
  segmentation.
\newblock In {\em Eur. Conf. Comput. Vis.}, pages 448--465. 2020.

\bibitem{Ronneberger2015UNet}
Olaf Ronneberger, Philipp Fischer, and Thomas Brox.
\newblock U-net: Convolutional networks for biomedical image segmentation.
\newblock In {\em Medical Image Computing and Computer Assisted Intervention},
  pages 234--241. 2015.

\bibitem{Roth2015DeepOrgan}
Holger~R. Roth, Le Lu, Amal Farag, Hoo{-}Chang Shin, Jiamin Liu, Evrim~B.
  Turkbey, and Ronald~M. Summers.
\newblock Deeporgan: Multi-level deep convolutional networks for automated
  pancreas segmentation.
\newblock In {\em Medical Image Computing and Computer Assisted Intervention},
  pages 556--564. 2015.

\bibitem{Roth2014LNDetection}
Holger~R. Roth, Le Lu, Ari Seff, Kevin~M. Cherry, Joanne Hoffman, Shijun Wang,
  Jiamin Liu, Evrim Turkbey, and Ronald~M. Summers.
\newblock A new 2.5d representation for lymph node detection using random sets
  of deep convolutional neural network observations.
\newblock In {\em Medical Image Computing and Computer Assisted Intervention},
  pages 520--527. 2014.

\bibitem{Shao2019Attentive}
Qingbin Shao, Lijun Gong, Kai Ma, Hualuo Liu, and Yefeng Zheng.
\newblock Attentive {CT} lesion detection using deep pyramid inference with
  multi-scale booster.
\newblock In {\em Medical Image Computing and Computer Assisted Intervention},
  pages 301--309. 2019.

\bibitem{Shin2016CNN}
Hoo{-}Chang Shin, Holger~R. Roth, Mingchen Gao, Le Lu, Ziyue Xu, Isabella
  Nogues, Jianhua Yao, Daniel~J. Mollura, and Ronald~M. Summers.
\newblock Deep convolutional neural networks for computer-aided detection:
  {CNN} architectures, dataset characteristics and transfer learning.
\newblock {\em IEEE Trans. Med. Imaging}, 35(5):1285--1298, 2016.

\bibitem{Tan2016OvarianCancer}
Maxine Tan, Zheng Li, Yuchen Qiu, Scott~D. McMeekin, Theresa~C. Thai, Kai Ding,
  Kathleen~N. Moore, Hong Liu, and Bin Zheng.
\newblock A new approach to evaluate drug treatment response of ovarian cancer
  patients based on deformable image registration.
\newblock {\em IEEE Trans. Med. Imaging}, 35(1):316--325, 2016.

\bibitem{tang2018semi}
Youbao Tang, Adam~P Harrison, Mohammadhadi Bagheri, Jing Xiao, and Ronald~M
  Summers.
\newblock Semi-automatic {RECIST} labeling on {CT} scans with cascaded
  convolutional neural networks.
\newblock In {\em Medical Image Computing and Computer Assisted Intervention},
  pages 405--413. 2018.

\bibitem{Tang2019Uldor}
Youbao Tang, Ke Yan, Yuxing Tang, Jiamin Liu, Jin Xiao, and Ronald~M. Summers.
\newblock Uldor: {A} universal lesion detector for ct scans with pseudo masks
  and hard negative example mining.
\newblock In {\em IEEE Int. Symposium on Biomedical Imaging}, pages 833--836.
  2019.

\bibitem{Tang2020OneClick}
Youbao Tang, Ke Yan, Jing Xiao, and Ronald~M. Summers.
\newblock One click lesion {RECIST} measurement and segmentation on {CT} scans.
\newblock In {\em Medical Image Computing and Computer Assisted Intervention},
  pages 573--583. 2020.

\bibitem{Tao2016SINT}
Ran Tao, Efstratios Gavves, and Arnold W.~M. Smeulders.
\newblock Siamese instance search for tracking.
\newblock In {\em IEEE Conf. Comput. Vis. Pattern Recog.}, pages 1420--1429.
  2016.

\bibitem{Valmadre2017CFNet}
Jack Valmadre, Luca Bertinetto, Jo{\~{a}}o~F. Henriques, Andrea Vedaldi, and
  Philip H.~S. Torr.
\newblock End-to-end representation learning for correlation filter based
  tracking.
\newblock In {\em IEEE Conf. Comput. Vis. Pattern Recog.}, pages 5000--5008.
  2017.

\bibitem{Wang2015FCNT}
Lijun Wang, Wanli Ouyang, Xiaogang Wang, and Huchuan Lu.
\newblock Visual tracking with fully convolutional networks.
\newblock In {\em Int. Conf. Comput. Vis.}, pages 3119--3127. 2015.

\bibitem{Wang2015SODLT}
Naiyan Wang, Siyi Li, Abhinav Gupta, and Dit{-}Yan Yeung.
\newblock Transferring rich feature hierarchies for robust visual tracking.
\newblock {\em CoRR}, abs/1501.04587, 2015.

\bibitem{Wang2019UOD}
Xudong Wang, Zhaowei Cai, Dashan Gao, and Nuno Vasconcelos.
\newblock Towards universal object detection by domain attention.
\newblock In {\em IEEE Conf. Comput. Vis. Pattern Recog.}, pages 7289--7298.
  2019.

\bibitem{Xie2017HED}
Saining Xie and Zhuowen Tu.
\newblock Holistically-nested edge detection.
\newblock {\em Int. J. Comput. Vis.}, 125(1-3):3--18, 2017.

\bibitem{Xu2016SixRegistration}
Zhoubing Xu, Christopher~P. Lee, Mattias~P. Heinrich, Marc Modat, Daniel
  Rueckert, S{\'{e}}bastien Ourselin, Richard~G. Abramson, and Bennett~A.
  Landman.
\newblock Evaluation of six registration methods for the human abdomen on
  clinically acquired {CT}.
\newblock {\em IEEE Trans. Biomed. Eng.}, 63(8):1563--1572, 2016.

\bibitem{Yan2020Lens}
Ke Yan, Jinzheng Cai, Youjing Zheng, Adam~P. Harrison, Dakai Jin, Youvao Tang,
  Yuxing Tang, Lingyun Huang, Jing Xiao, and Le Lu.
\newblock Learning from multiple datasets with heterogeneous and partial labels
  for universal lesion detection in {CT}.
\newblock {\em CoRR}, abs/2009.02577, 2020.

\bibitem{Yan2019LesaNet}
Ke Yan, Yifan Peng, Veit Sandfort, Mohammadhadi Bagheri, Zhiyong Lu, and
  Ronald~M. Summers.
\newblock Holistic and comprehensive annotation of clinically significant
  findings on diverse {CT} images: Learning from radiology reports and label
  ontology.
\newblock In {\em IEEE Conf. Comput. Vis. Pattern Recog.}, pages 8523--8532.
  2019.

\bibitem{Yan2019Mulan}
Ke Yan, Youbao Tang, Yifan Peng, Veit Sandfort, Mohammadhadi Bagheri, Zhiyong
  Lu, and Ronald~M. Summers.
\newblock {MULAN:} multitask universal lesion analysis network for joint lesion
  detection, tagging, and segmentation.
\newblock In {\em Medical Image Computing and Computer Assisted Intervention},
  pages 194--202. 2019.

\bibitem{Yan2018DeepLesion}
Ke Yan, Xiaosong Wang, Le Lu, and Ronald~M. Summers.
\newblock Deeplesion: automated mining of large-scale lesion annotations and
  universal lesion detection with deep learning.
\newblock {\em J. Med. Imaging}, 5(3), 2018.

\bibitem{Yan2018LesionGraph}
Ke Yan, Xiaosong Wang, Le Lu, Ling Zhang, Adam~P. Harrison, Mohammadhadi
  Bagheri, and Ronald~M. Summers.
\newblock Deep lesion graphs in the wild: Relationship learning and
  organization of significant radiology image findings in a diverse large-scale
  lesion database.
\newblock In {\em IEEE Conf. Comput. Vis. Pattern Recog.}, pages 9261--9270.
  2018.

\bibitem{Yao2013PBV}
Rui Yao, Qinfeng Shi, Chunhua Shen, Yanning Zhang, and Anton van~den Hengel.
\newblock Part-based visual tracking with online latent structural learning.
\newblock In {\em IEEE Conf. Comput. Vis. Pattern Recog.}, pages 2363--2370.
  2013.

\bibitem{Zhang2020PDAC}
Ling Zhang, Yu Shi, Jiawen Yao, Yun Bian, Kai Cao, Dakai Jin, Jing Xiao, and Le
  Lu.
\newblock Robust pancreatic ductal adenocarcinoma segmentation with
  multi-institutional multi-phase partially-annotated {CT} scans.
\newblock In {\em Medical Image Computing and Computer Assisted Intervention},
  pages 491--500. 2020.

\bibitem{Zhou2019CenterNet}
Xingyi Zhou, Dequan Wang, and Philipp Kr{\"{a}}henb{\"{u}}hl.
\newblock Objects as points.
\newblock {\em CoRR}, abs/1904.07850, 2019.

\end{thebibliography}
}

\newpage

\section*{Supplementary Materials}

{\bf Parameter analysis with more details.} Due to constraints of space, we have omit some details of experiments that reported in Table. \textcolor{red}{3} in the main manuscript. As promised, we show the complete version here in \Table{pa}.

\begin{figure*}[t!]
\centering
\begin{tabular}{rl}
  \begin{minipage}{.49\textwidth}
      \includegraphics[width=.98\linewidth]{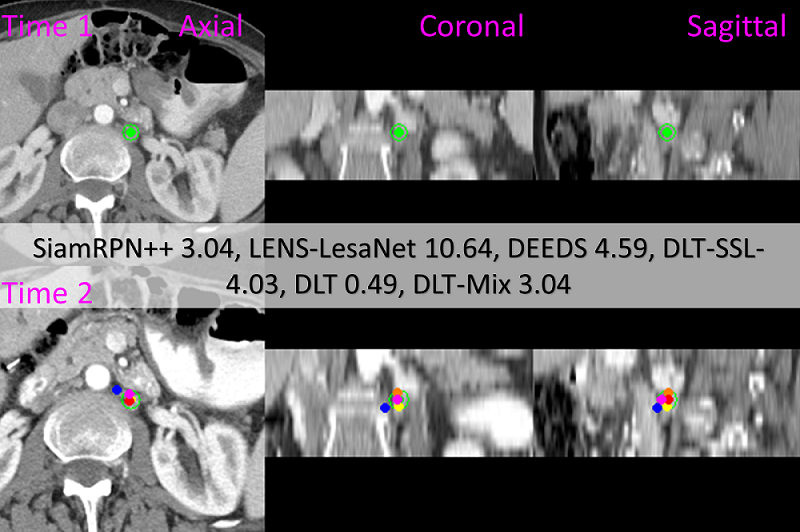}
  \end{minipage}
  &
  \begin{minipage}{.49\textwidth}
      \includegraphics[width=.98\linewidth]{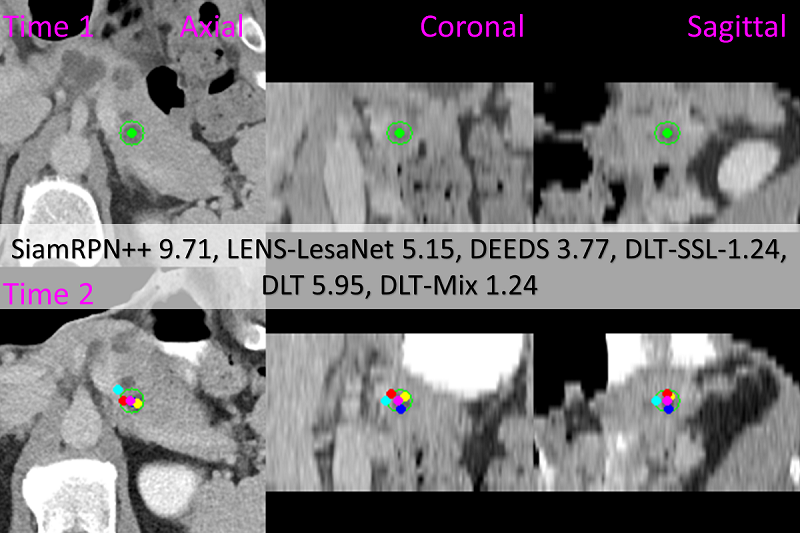}
  \end{minipage}
  \\
  \begin{minipage}{.49\textwidth}
      \includegraphics[width=.98\linewidth, trim=0.1in 0.025in 0.1in 0.105in, clip]{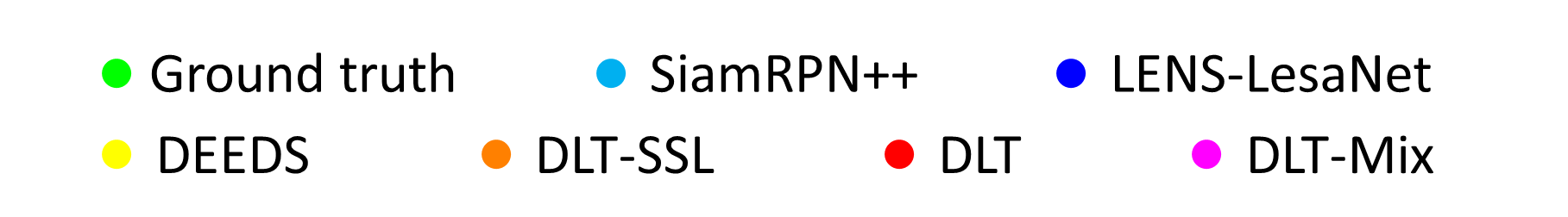}
  \end{minipage}
  &
  \begin{minipage}{.49\textwidth}
      \includegraphics[width=.98\linewidth, trim=0.1in 0.025in 0.1in 0.105in, clip]{figures/DLT-Examples/legend.png}
  \end{minipage}
  \\
  \begin{minipage}{.49\textwidth}
      \includegraphics[width=.98\linewidth]{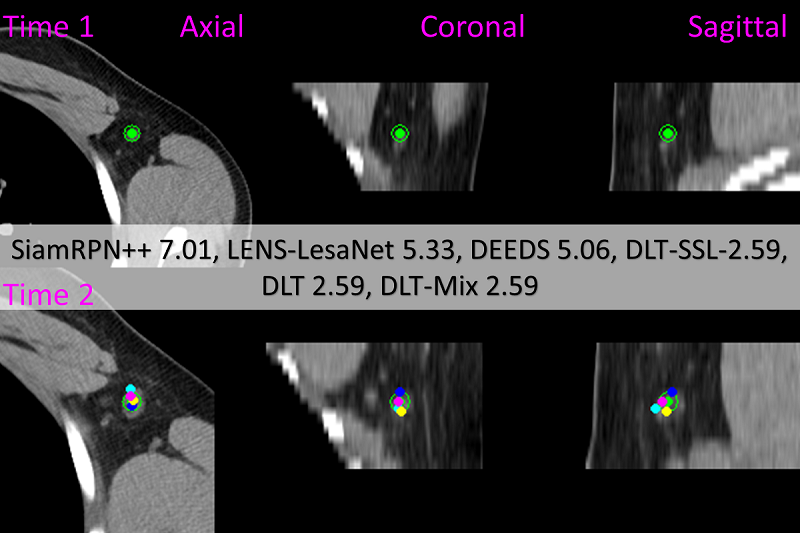}
  \end{minipage}
  &
  \begin{minipage}{.49\textwidth}
      \includegraphics[width=.98\linewidth]{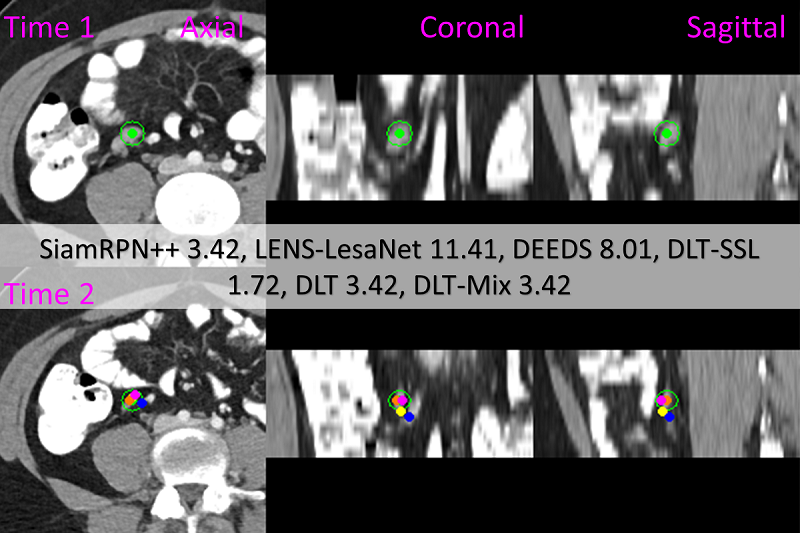}
  \end{minipage}
  \\
  \begin{minipage}{.49\textwidth}
      \includegraphics[width=.98\linewidth, trim=0.1in 0.025in 0.1in 0.105in, clip]{figures/DLT-Examples/legend.png}
  \end{minipage}
  &
  \begin{minipage}{.49\textwidth}
      \includegraphics[width=.98\linewidth, trim=0.1in 0.025in 0.1in 0.105in, clip]{figures/DLT-Examples/legend.png}
  \end{minipage}
\end{tabular}
\caption{Comparison of our methods, \ie{}, \ac{my_method}, \ac{my_method}-\ac{SSL}, \ac{my_method}-Mix, with three state-of-the-art trackers including a Siamese networks based tracker -- SiamRPN++, a leading registration algorithm -- DEEDS, and a detector based tracker -- LENS-LesaNet. Offsets from the predicted lesion centers to the manually labeled center are reported in $mm$.}
\label{fig:compr1}
\end{figure*}

\begin{figure*}[t!]
\centering
\begin{tabular}{rl}
  \begin{minipage}{.49\textwidth}
      \includegraphics[width=.98\linewidth]{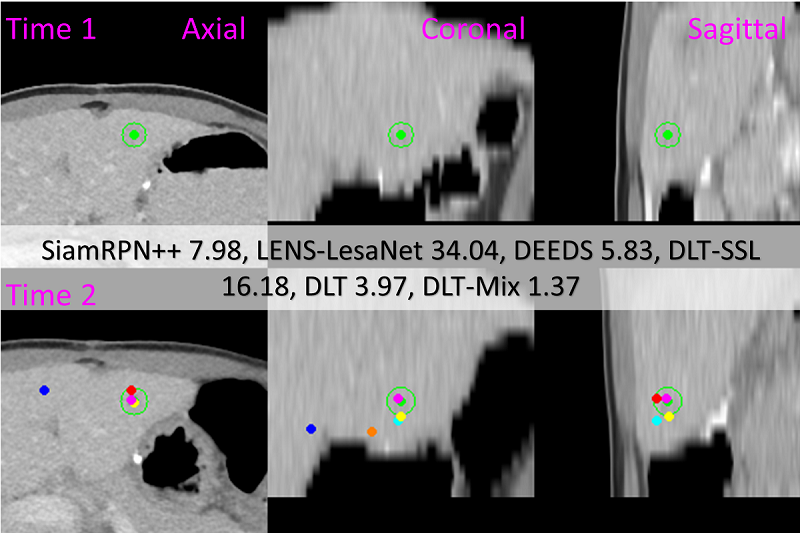}
  \end{minipage}
  &
  \begin{minipage}{.49\textwidth}
      \includegraphics[width=.98\linewidth]{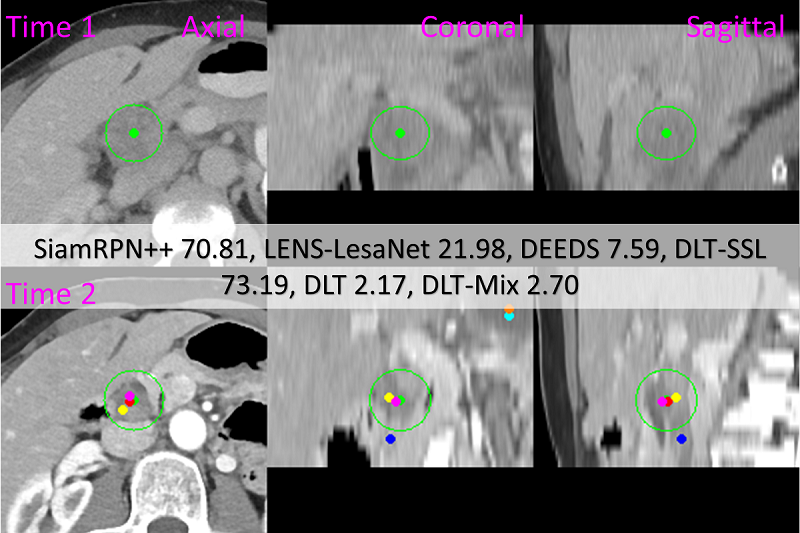}
  \end{minipage}
  \\
  \begin{minipage}{.49\textwidth}
      \includegraphics[width=.98\linewidth, trim=0.1in 0.025in 0.1in 0.105in, clip]{figures/DLT-Examples/legend.png}
  \end{minipage}
  &
  \begin{minipage}{.49\textwidth}
      \includegraphics[width=.98\linewidth, trim=0.1in 0.025in 0.1in 0.105in, clip]{figures/DLT-Examples/legend.png}
  \end{minipage}
  \\
  \begin{minipage}{.49\textwidth}
      \includegraphics[width=.98\linewidth]{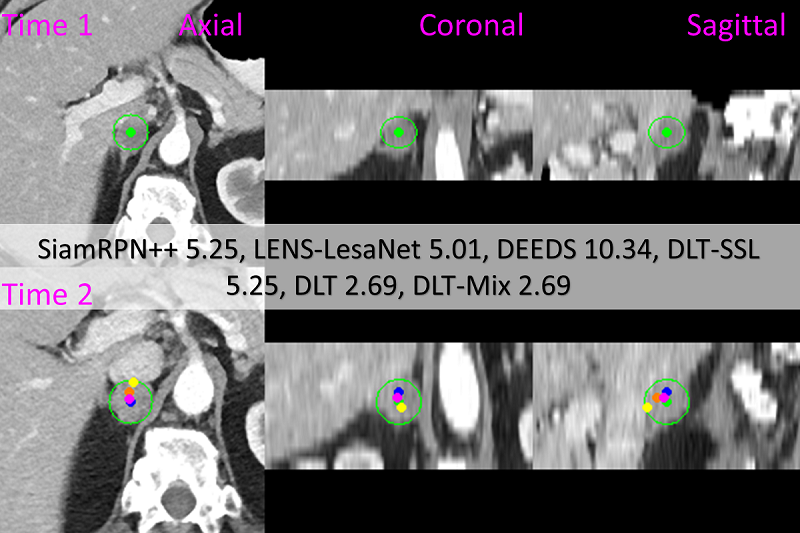}
  \end{minipage}
  &
  \begin{minipage}{.49\textwidth}
      \includegraphics[width=.98\linewidth]{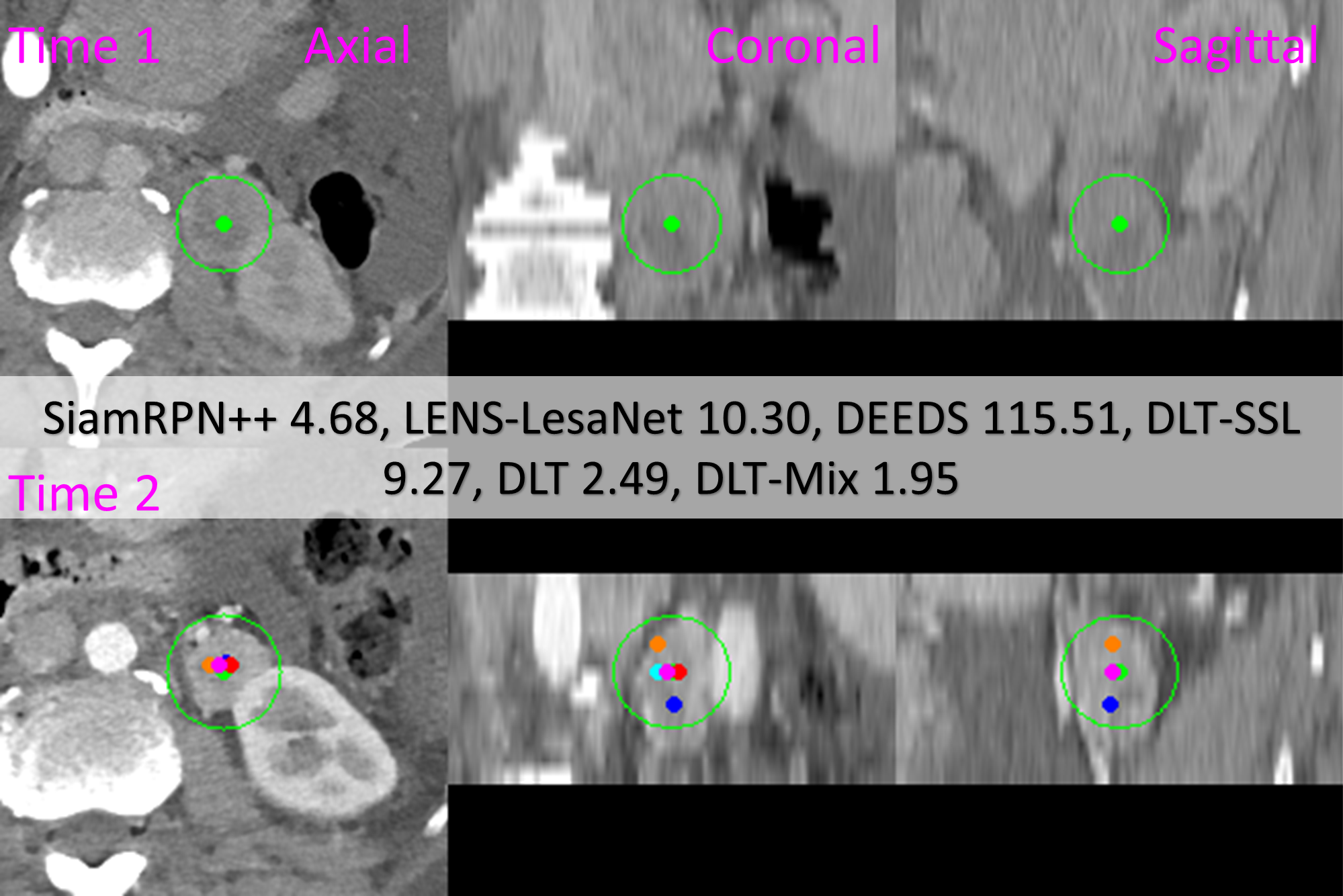}
  \end{minipage}
  \\
  \begin{minipage}{.49\textwidth}
      \includegraphics[width=.98\linewidth, trim=0.1in 0.025in 0.1in 0.105in, clip]{figures/DLT-Examples/legend.png}
  \end{minipage}
  &
  \begin{minipage}{.49\textwidth}
      \includegraphics[width=.98\linewidth, trim=0.1in 0.025in 0.1in 0.105in, clip]{figures/DLT-Examples/legend.png}
  \end{minipage}
  \\
  \begin{minipage}{.49\textwidth}
      \includegraphics[width=1\linewidth]{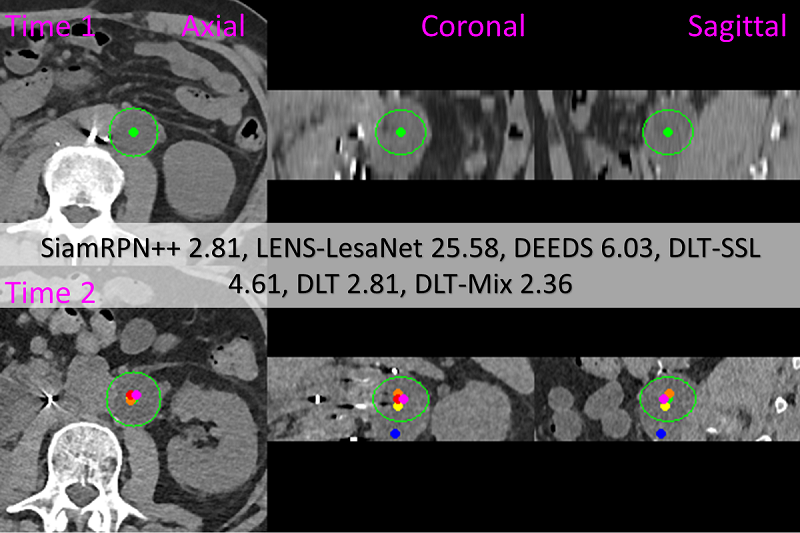}
  \end{minipage}
  &
  \begin{minipage}{.49\textwidth}
      \includegraphics[width=1\linewidth]{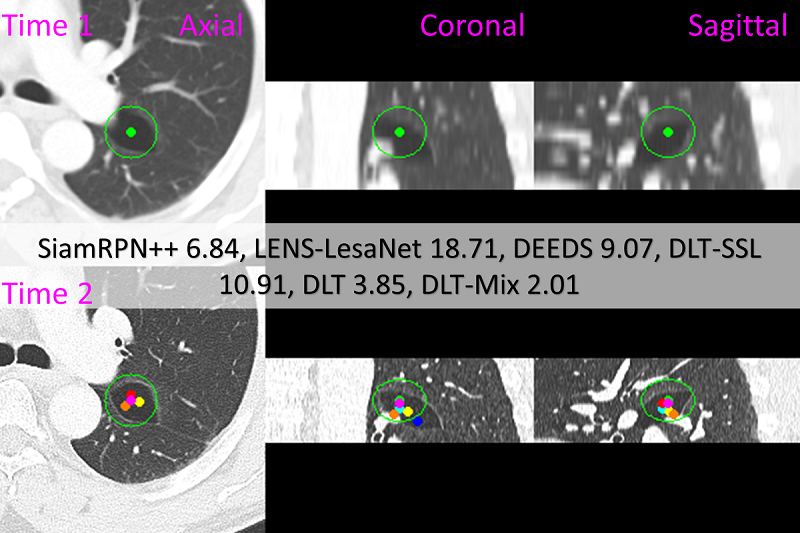}
  \end{minipage}
  \\
  \begin{minipage}{.49\textwidth}
      \includegraphics[width=1\linewidth, trim=0.1in 0.025in 0.1in 0.105in, clip]{figures/DLT-Examples/legend.png}
  \end{minipage}
  &
  \begin{minipage}{.49\textwidth}
      \includegraphics[width=1\linewidth, trim=0.1in 0.025in 0.1in 0.105in, clip]{figures/DLT-Examples/legend.png}
  \end{minipage}
\end{tabular}
\caption{comparison of our methods, \ie{}, \ac{my_method}, \ac{my_method}-\ac{SSL}, \ac{my_method}-Mix, with three state-of-the-art trackers including a Siamese networks based tracker -- SiamRPN++, a leading registration algorithm -- DEEDS, and a detector based tracker -- LENS-LesaNet. Offsets from the predicted lesion centers to the manually labeled center are reported in $mm$.}
\label{fig:compr2}
\end{figure*}

{\bf More visualization examples for method comparison.} In Fig. \textcolor{red}{4} of the main manuscript, we compared our methods with three state-of-the-art trackers. Here, we show more examples in \Fig{compr1} and \Fig{compr2}. All case are shown with representative axial, coronal, and sagittal slices to accurately illustrate 3D locations.  

For 2D visualization, we orthographically projected the lesion center from 3D. These centers were projected from any axial slices within 10$mm$ of the ground truth axial slices (most CTs have 5$mm$ slice thickness). Thus, in the second example of Fig.~\textcolor{red}{4} in the main manuscript, DEEDS is actually located further away in the $z$ direction, despite the visual appearance. Some centers in samples 5 and 7 are invisible because they overlap and/or they are located outside of the +/- 10$mm$ limit

\begin{figure*}[t!]
\centering
\begin{tabular}{c}
  \begin{minipage}{.98\textwidth}
    \includegraphics[width=.98\linewidth]{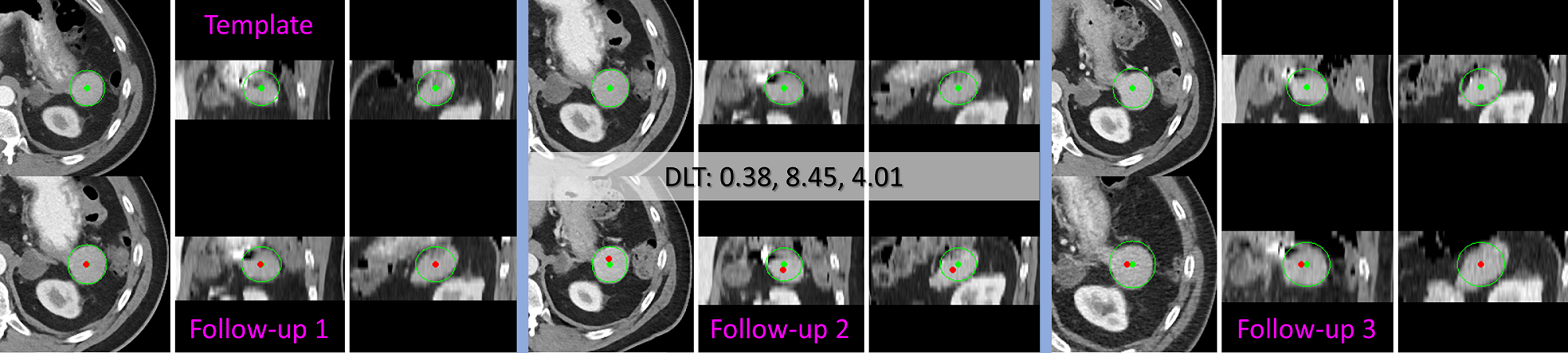}
  \end{minipage}
  \vspace{2mm}
  \\
  \begin{minipage}{.98\textwidth}
    \includegraphics[width=.98\linewidth]{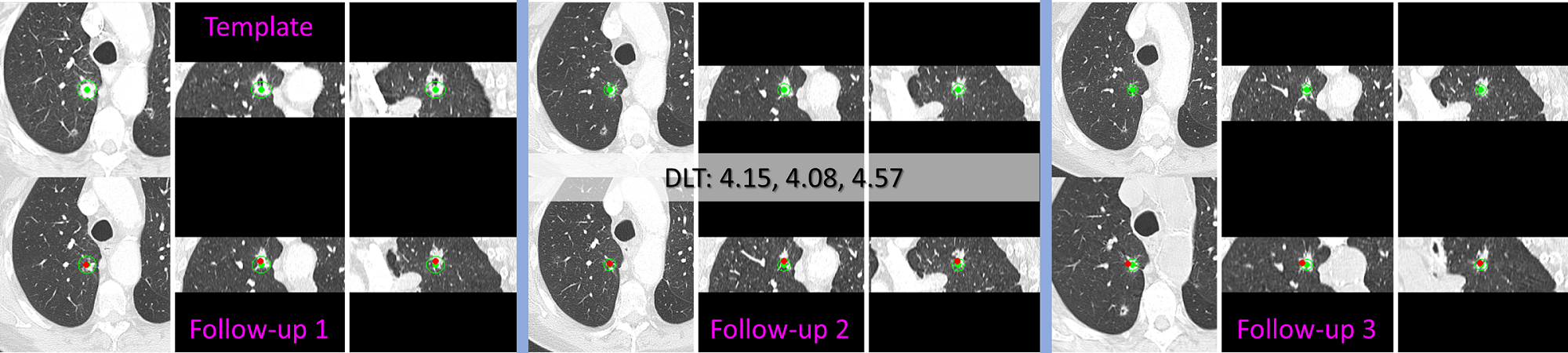}
  \end{minipage}
  \vspace{2mm}
  \\
  \begin{minipage}{.98\textwidth}
    \includegraphics[width=.98\linewidth]{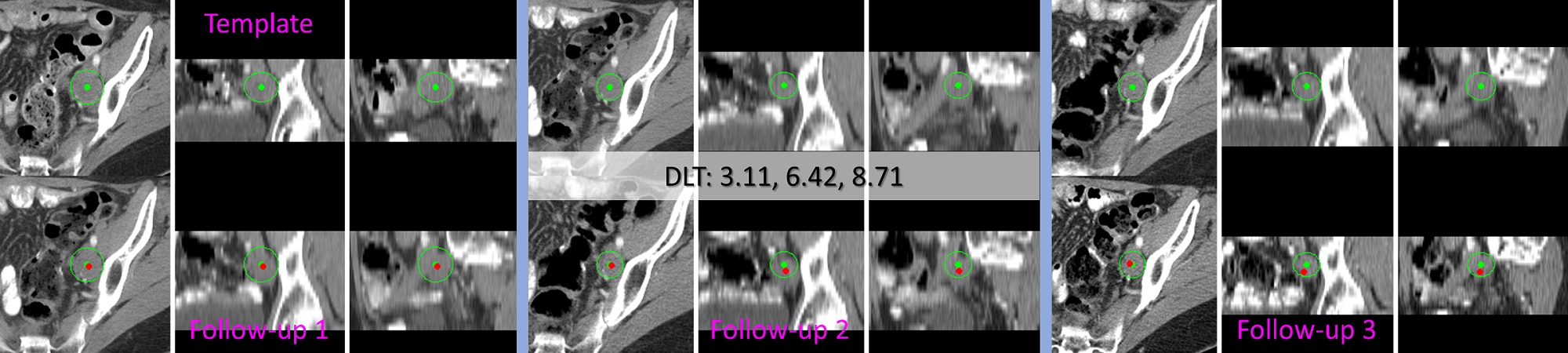}
  \end{minipage}
  \vspace{2mm}
  \\
  \begin{minipage}{.98\textwidth}
    \includegraphics[width=.98\linewidth]{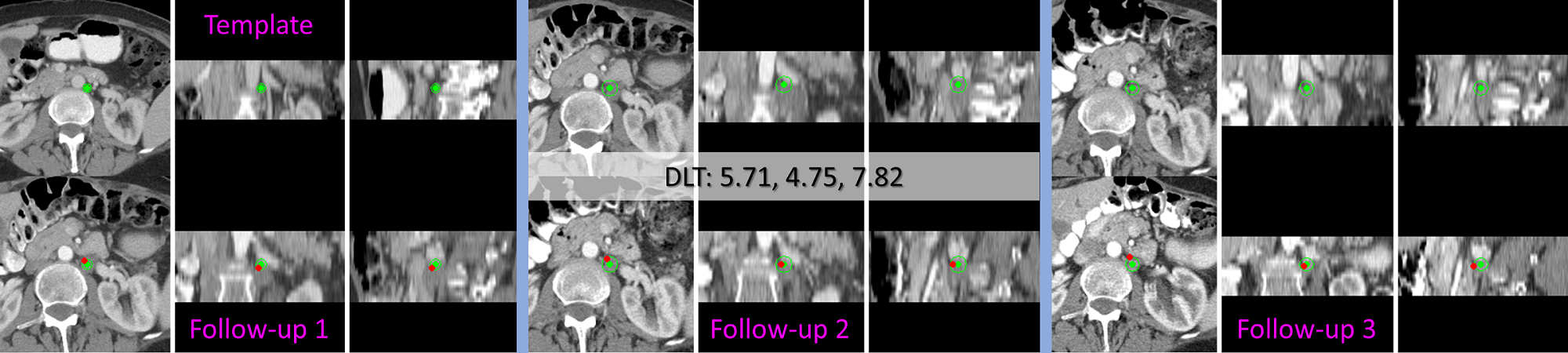}
  \end{minipage}
  \vspace{2mm}
  \\
  \begin{minipage}{.98\textwidth}
    \includegraphics[width=.98\linewidth]{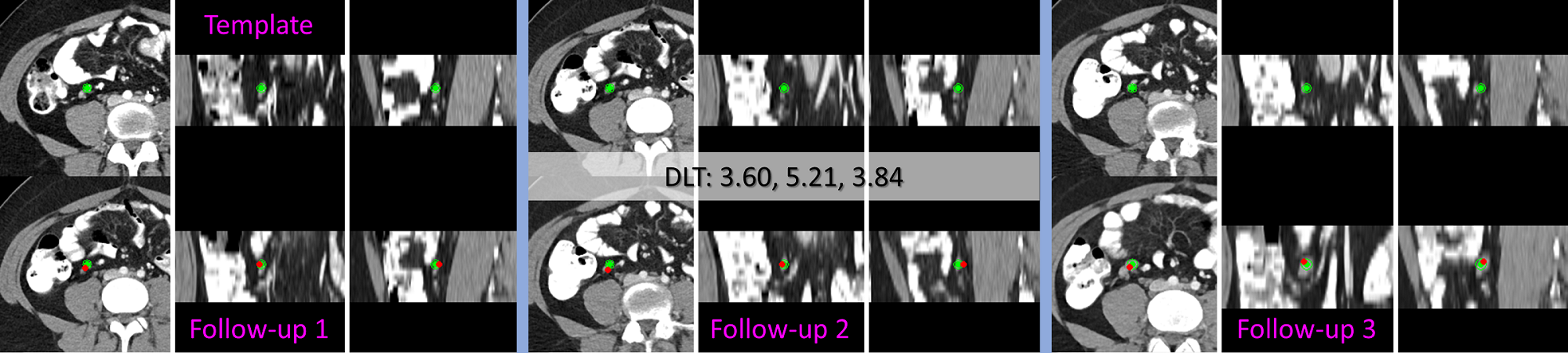}
  \end{minipage}
  \vspace{2mm}
\end{tabular}
\caption{Lesion tracking through three follow ups using the proposed \ac{my_method}. The template image is sampled from the first exam, and then follow-up 1, 2, and 3 are sampled from times of the second, third, and fourth exams, respectively. Green and red points represent the manually labeled and \ac{my_method} predicted centers, respectively. Only the lesion center and radius at the first time point is given. Offsets from the \ac{my_method} predicted lesion center to the manually labeled center are reported in $mm$.}
\label{fig:4d1}
\end{figure*}

\begin{figure*}[t!]
\centering
\begin{tabular}{c}
  \begin{minipage}{.98\textwidth}
    \includegraphics[width=.98\linewidth]{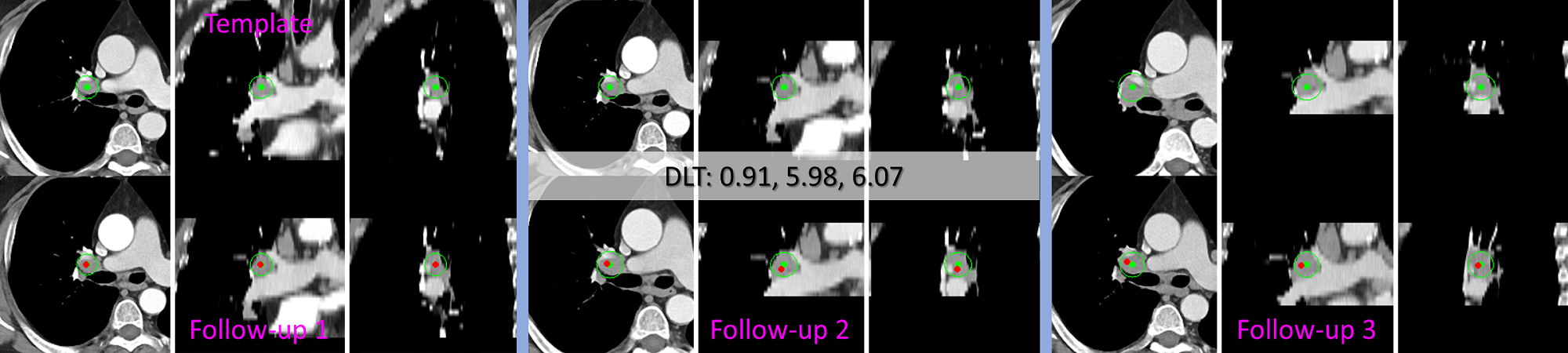}
  \end{minipage}
  \vspace{2mm}
  \\
  \begin{minipage}{.98\textwidth}
    \includegraphics[width=.98\linewidth]{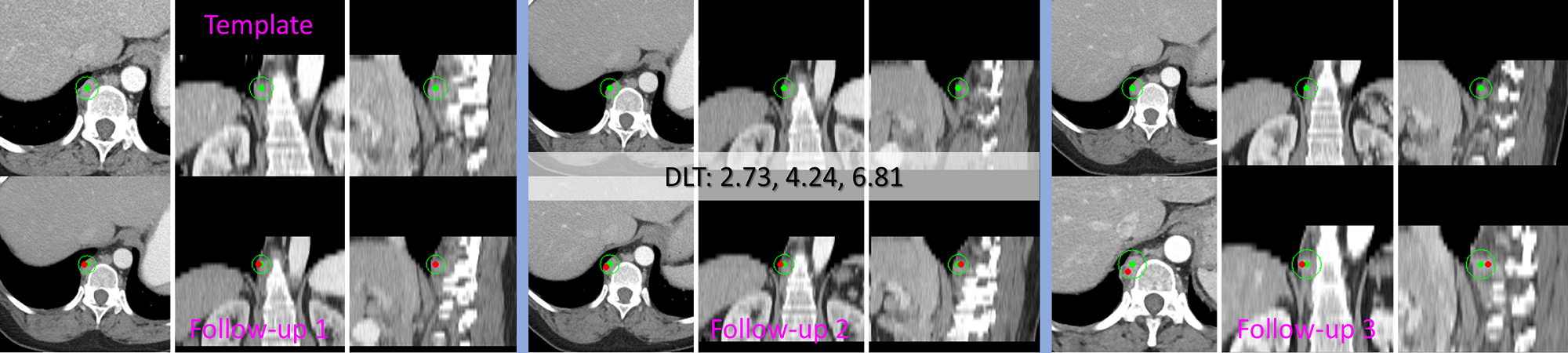}
  \end{minipage}
  \vspace{2mm}
  \\
  \begin{minipage}{.98\textwidth}
    \includegraphics[width=.98\linewidth]{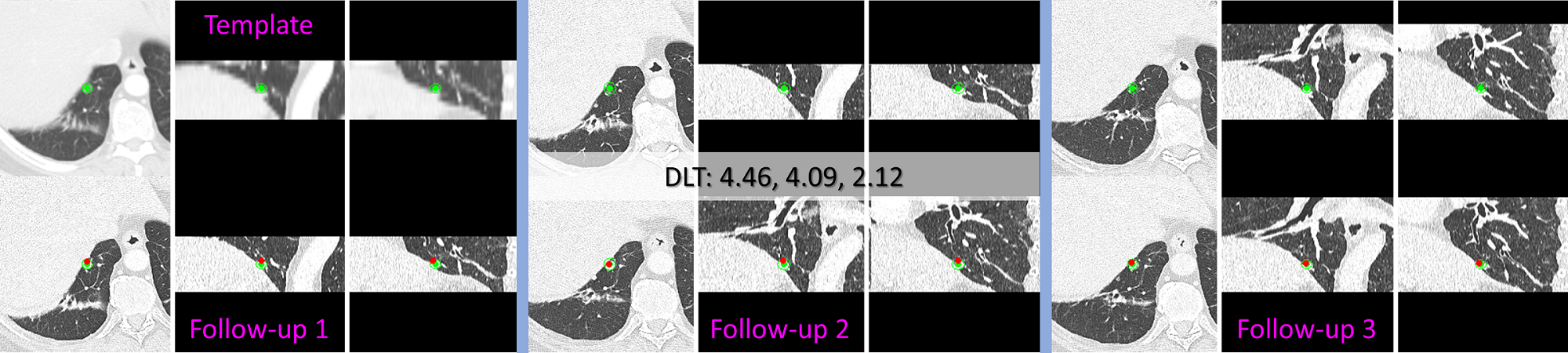}
  \end{minipage}
  \vspace{2mm}
  \\
  \begin{minipage}{.98\textwidth}
    \includegraphics[width=.98\linewidth]{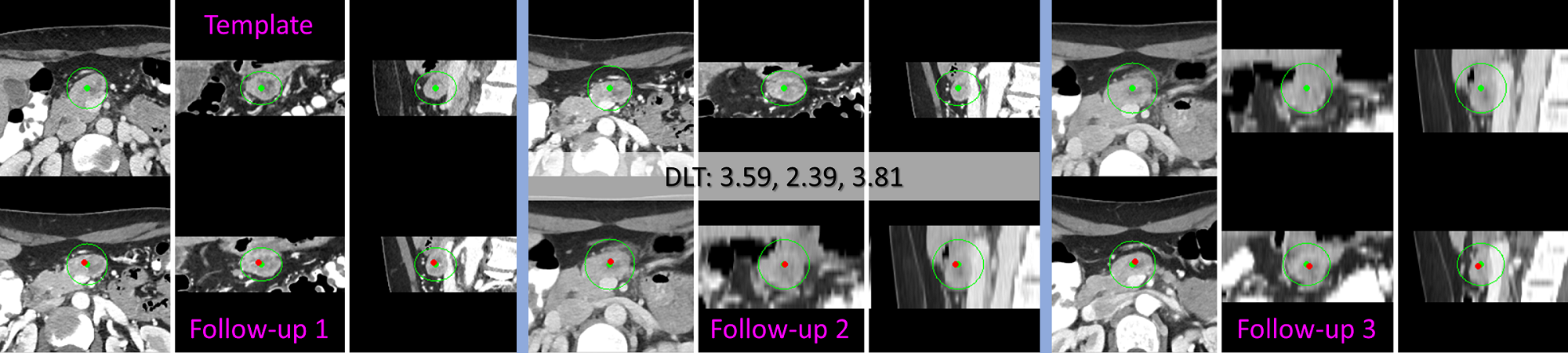}
  \end{minipage}
  \vspace{2mm}
  \\
  \begin{minipage}{.98\textwidth}
    \includegraphics[width=.98\linewidth]{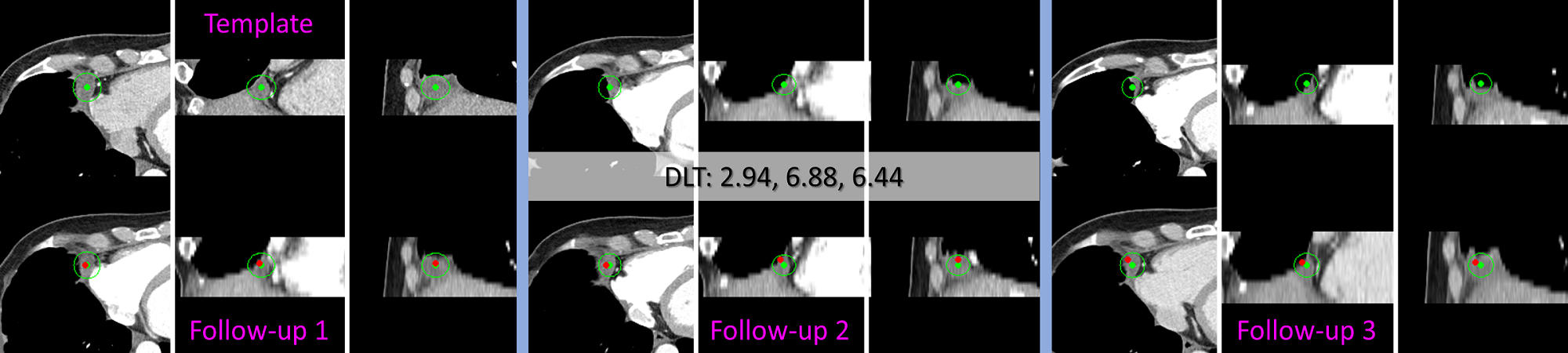}
  \end{minipage}
  \vspace{2mm}
\end{tabular}
\caption{Lesion tracking through three follow ups using the proposed \ac{my_method}. Green and red points present the manual labeled and \ac{my_method} predicted centers, respectively. Only the lesion center and radius at the first time point is given. Offsets from the \ac{my_method} predicted lesion center to the manual labeled center are reported in $mm$.}
\label{fig:4d2}
\end{figure*}

\begin{figure*}[t!]
\centering
\begin{tabular}{c}
  \begin{minipage}{.98\textwidth}
    \includegraphics[width=.98\linewidth]{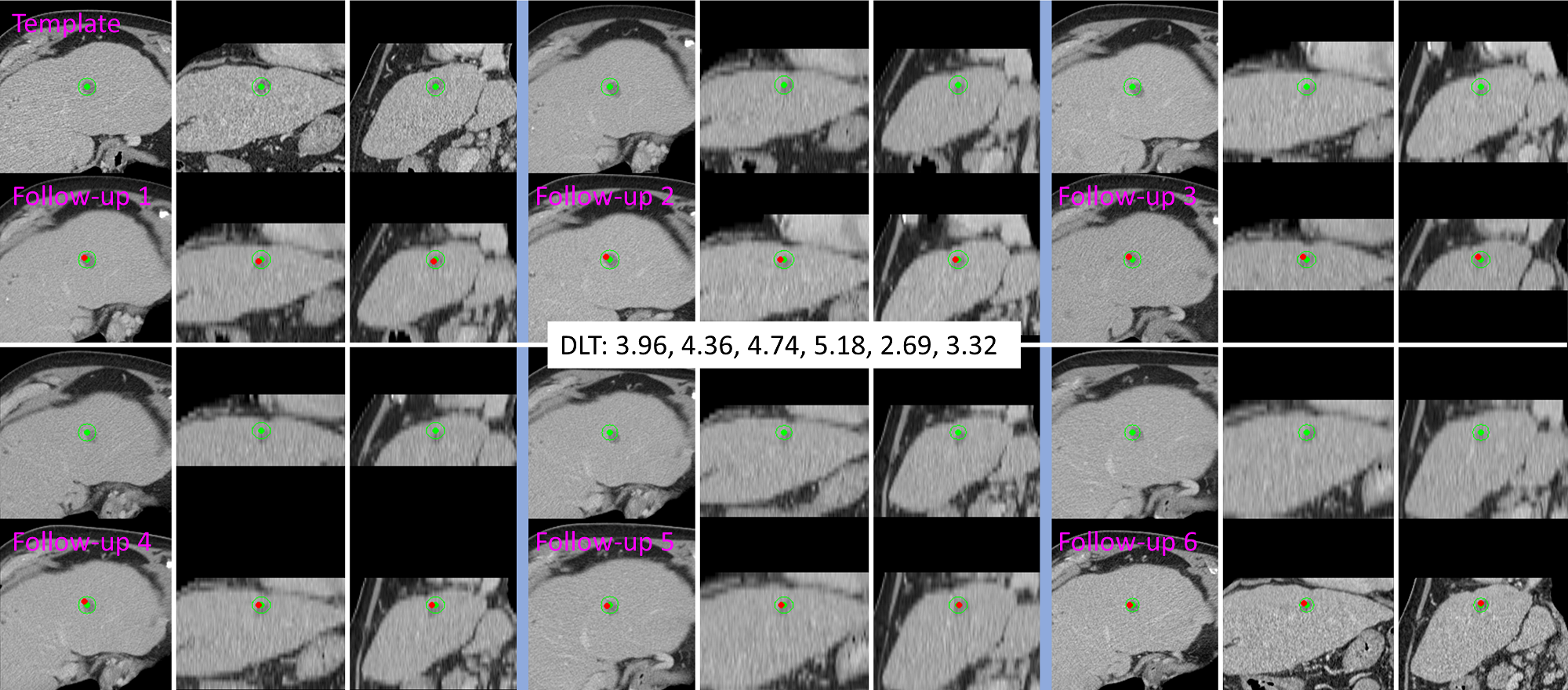}
  \end{minipage}
  \\
  \begin{minipage}{.98\textwidth}
    \includegraphics[width=.98\linewidth]{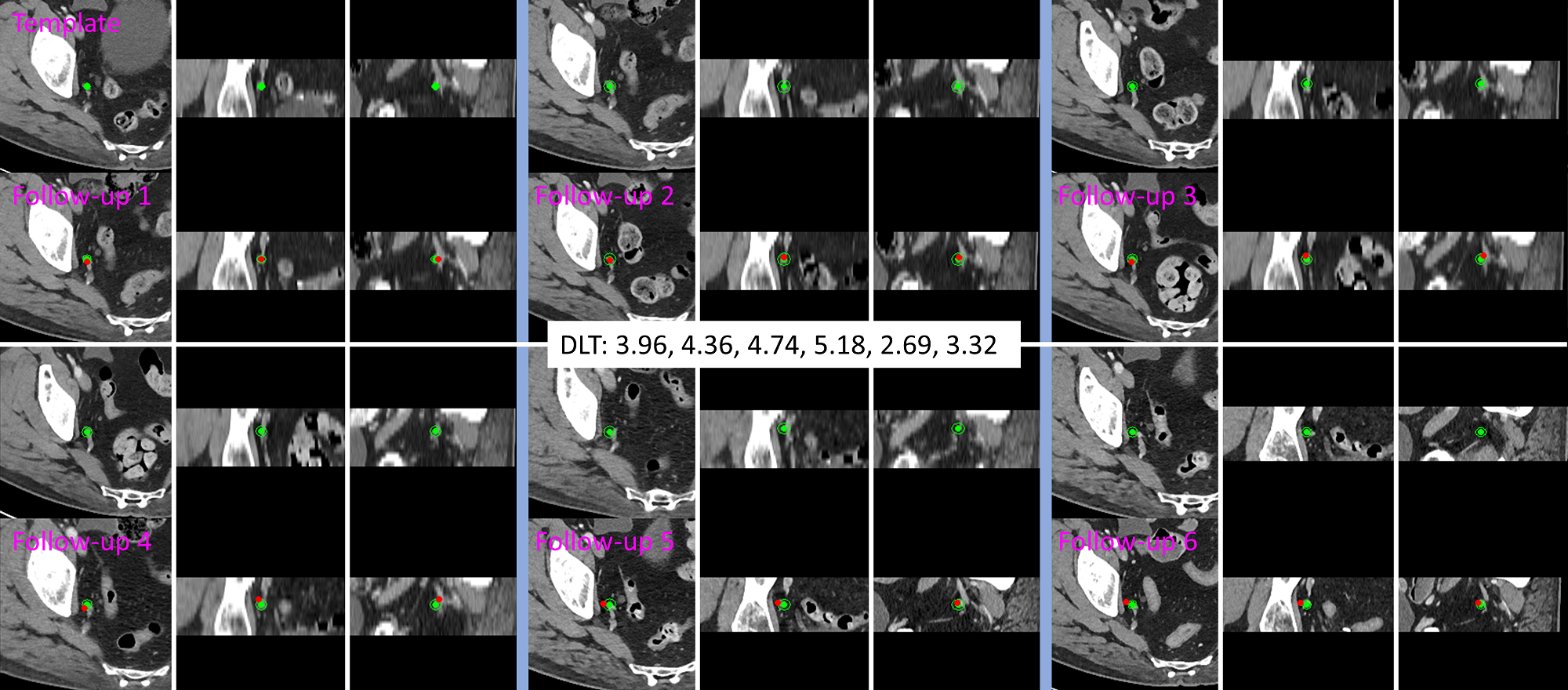}
  \end{minipage}
  \\
\end{tabular}
\caption{Lesion tracking through six follow ups using the proposed \ac{my_method}. Green and red points represent the manual labeled and \ac{my_method} predicted centers, respectively. Only the lesion center and radius at the first time point is given. Offsets from the \ac{my_method} predicted lesion center to the manual labeled center are reported in $mm$.}
\label{fig:6d1}
\end{figure*}

{\bf Visualization examples for lesion tracking with multiple follow ups.} We show lesion tracking using \ac{my_method} with three follow-ups in \Fig{4d1} and \Fig{4d2}. In \Fig{6d1}, we show \ac{my_method} tracks lesions up to six follow-ups. For lesion tracking with multiple follow-ups, \ac{my_method} is only provided with the location of the target lesion in the initial template image. 

{\bf Robustness analysis with more details.} Due to constrains of space, we have omit ``CPM@Radius'' in the Table~\textcolor{red}{2} of the main manuscript and kept ``CPM@10$mm$'' as it is a more tighter evaluation. In \Table{robustness}, measurements of performance of different trackers delivered by ``CPM@Radius'' follow the same trend as the results measured by ``CPM@10$mm$''. With ``CPM@Radius'', DLT-Mix remains to be the best approach. DEEDS is the most vulnerable method with over 10\% drop in ``CPM@Radius''. In comparison, DLT-Mix only drops 1.87\%.

\begin{table*}[t!]
    \small
    \centering
    \begin{tabular}{l|c|cc|c|c|cc|ccc}
    \hline
    Model & Ablation study & \multicolumn{2}{c|}{Eq. \textcolor{red}{6}: $K_g$} & $\psi, \phi$ & Eq. \textcolor{red}{3} & \multicolumn{2}{c|}{Eq. \textcolor{red}{2}: $G$} & Valid      & Test       & Speed \\
    id    &                & size & learn                                       & dim.         & fusion                 & size     & $\Delta\mu_t$                         & MED ($mm$) & MED ($mm$) & spv   \\
    \hline
    $a$ & w/o $K_g$             & NA      & NA             & 64 & multiply & 4$r$ &\XSolidBrush   & 8.77$\pm$9.88 ($\uparrow$1.69) & 9.29$\pm$10.2 & 1.44 \\
    $b$ & smaller $K_g$         & 7,7,7   & \CheckmarkBold & 64 & multiply & 4$r$ &\XSolidBrush   & 8.26$\pm$9.40 ($\uparrow$1.18) & 9.41$\pm$10.2 & 2.38 \\
    $c$ & greater $K_g$         & 7,15,15 & \CheckmarkBold & 64 & multiply & 4$r$ &\XSolidBrush   & 7.24$\pm$5.64 ($\uparrow$0.16) & 7.67$\pm$8.78 & 24.1 \\
    $d$ & smaller $G_t, G_s$    & 7,11,11 & \CheckmarkBold & 64 & multiply & 2$r$ &\XSolidBrush   & 7.56$\pm$8.95 ($\uparrow$0.48) & 7.51$\pm$8.39 & 3.51 \\
    $e$ & greater $G_t, G_s$    & 7,11,11 & \CheckmarkBold & 64 & multiply & 8$r$ &\XSolidBrush   & 8.40$\pm$9.23 ($\uparrow$1.32) & 8.81$\pm$9.80 & 3.51 \\ 
    $f$ & smaller feat. dim.    & 7,11,11 & \CheckmarkBold & 32 & multiply & 4$r$ &\XSolidBrush   & 7.23$\pm$6.17 ($\uparrow$0.15) & 8.72$\pm$16.6 & 2.25 \\
    $g$ & greater feat. dim.    & 7,11,11 & \CheckmarkBold &128 & multiply & 4$r$ &\XSolidBrush   & 7.15$\pm$6.99 ($\uparrow$0.07) & 7.91$\pm$9.29 & 5.83 \\
    $h$ & w/o \acs{anatomy_net} & 7,11,11 & \CheckmarkBold & 64 & NA       & NA   & NA            & 8.23$\pm$9.44 ($\uparrow$1.15) & 9.34$\pm$10.0 & 3.51 \\
    $i$ & w/o learn $K_g$       & 7,11,11 & \XSolidBrush   & 64 & multiply & 4$r$ &\XSolidBrush   & 7.61$\pm$9.02 ($\uparrow$0.53) & 7.98$\pm$9.26 & 3.51 \\ 
    & {\bf comparison baseline}           & 7,11,11 & \CheckmarkBold & 64 & multiply & 4$r$ &\XSolidBrush   & 7.08$\pm$5.25 ($\uparrow$0.00) & 7.95$\pm$8.96 & 3.51 \\ 
    \hline \hline
    & Eq. \textcolor{red}{3} with concat. & 7,11,11 & \CheckmarkBold & 64 & concat.  & 4$r$ &\CheckmarkBold & 6.85$\pm$9.47 ($\downarrow$0.23) & 7.94$\pm$9.22 & 5.91 \\
    & {\bf final configuration}       & 7,11,11 & \CheckmarkBold & 64 & multiply & 4$r$ &\CheckmarkBold & 6.69$\pm$5.62 ($\downarrow$0.39) & 6.98$\pm$8.95 & 3.51 \\
    \hline
    \end{tabular}
    \caption{Parameter analysis and ablation study of the proposed components.}
    \label{tbl:pa}
\end{table*}

\begin{table*}[t!]
   \small
   \centering 
   \begin{tabular}{lccc}
      \hline 
      Method & CPM@Radius & CPM@10$mm$ & MED ($mm$) \\
      \hline
      SiamRPN++ \cite{Li2019SiamRPN++}      & 71.52 ($\downarrow$ 8.79) & 51.27 ($\downarrow$ 17.6) & 10.6$\pm$10.3 ($\uparrow$ 2.3) \\
      DEEDS \cite{Heinrich2013DeedsBCV}     & 74.82 ($\downarrow$ 10.7) & 53.85 ($\downarrow$ 18.0) & 9.8$\pm$8.9 ($\uparrow$ 2.4) \\
      \ac{my_method}-SSL                    & 78.38 ($\downarrow$ 3.14) & 64.24 ($\downarrow$ 6.80) & 10.0$\pm$11.4 ($\uparrow$ 1.2) \\
      \ac{my_method}                        & 83.18 ($\downarrow$ 3.70) & 70.36 ($\downarrow$ 8.49) & 8.1$\pm$8.7 ($\uparrow$ 1.2) \\
      \ac{my_method}-Mix                    & {\bf 86.88 ($\downarrow$ 1.87)} & {\bf 75.03 ($\downarrow$ 3.62)} & {\bf 8.0$\pm$10.5 ($\uparrow$ 0.9)} \\
      \hline
   \end{tabular}
   \caption{Robustness evaluation. $\downarrow$ and $\uparrow$ demonstrate decrease and increase of measurements, respectively, compared with the values reported in Table~\textcolor{red}{1} in the main script.}
   \label{tbl:robustness}
\end{table*}

\end{document}